\documentclass[12pt,onecolumn]{IEEEtran}

\usepackage{float}
\usepackage{amsthm,amsmath,amsfonts,amssymb}
\usepackage{bm}
\usepackage{xspace,exscale,relsize}
\usepackage{fancybox,shadow}
\usepackage{graphicx}

\usepackage{xcolor}
\usepackage{extarrows}
\usepackage{comment}
\usepackage{booktabs} 
\usepackage{makecell, multirow}
\usepackage[noadjust]{cite}

\usepackage{mathrsfs}
\makeatletter
\let\over=\@@over \let\overwithdelims=\@@overwithdelims
\let\atop=\@@atop \let\atopwithdelims=\@@atopwithdelims
\let\above=\@@above \let\abovewithdelims=\@@abovewithdelims
\makeatother
\usepackage{rotating}

\usepackage{hyphenat}

\usepackage{ifpdf}

\usepackage{psfrag}

\usepackage{prettyref}

\usepackage{tikz}
\usetikzlibrary{arrows}
\tikzstyle{int}=[draw, fill=blue!20, minimum size=2em]
\tikzstyle{dot}=[circle, draw, fill=blue!20, minimum size=2em]
\tikzstyle{init} = [pin edge={to-,thin,black}]

\usepackage[
CJKbookmarks=true,
bookmarksnumbered=true,
bookmarksopen=true,
colorlinks=true,
citecolor=red,
linkcolor=blue,
anchorcolor=red,
urlcolor=blue
]{hyperref}

\usepackage[all]{xy}

\usepackage{paralist}
\usepackage{enumerate,enumitem}

\usepackage{algorithm}
\usepackage{algpseudocode}

\usepackage{mathtools}

\usepackage{mathtools}

\def\ba{{\boldsymbol a}}     
     
     \def\CC{\mathbb{C}}
\def\bd{{\boldsymbol d}}     
     \def\EE{\mathbb{E}}

   \def\bI{\boldsymbol I}

     \def\PP{\mathbb{P}}

\def\bu{{\boldsymbol u}}     
     
\def\bw{\boldsymbol w}   \def\bW{\boldsymbol W}  
\def\bx{\boldsymbol x}   \def\bX{\boldsymbol X}

\def\11{\mathbbm{1}}

\def\calC{{\cal  C}} 
 
\def\calE{{\cal  E}}

\def\calN{{\cal  N}} \def\cN{{\cal  N}}

\def\calR{{\cal  R}} 
\def\calS{{\cal  S}}

\newcommand{\bfsym}[1]{\ensuremath{\boldsymbol{#1}}}

\def\btheta{{\bfsym {\theta}}}

\newcommand{\Tr}{{\rm Tr}}

\usepackage{ifthen}

\ifx\eqref\undefined
\newcommand{\eqref}[1]{~(\ref{#1})}
\fi
\ifx\mod\undefined
\def\mod{\mathop{\rm mod}}
\fi

\usepackage{bm}

\def\argmin{\mathop{\rm argmin}}

\def\exp{\mathop{\rm exp}}

\def\EE{\Expect}

\def\PP{\mathbb{P}}

\def\simiid{\stackrel{iid}{\sim}}

\newcommand{\abs}[1]{\left| #1 \right|}

\makeatletter
\def\bbordermatrix#1{\begingroup \m@th
	\@tempdima 4.75\p@
	\setbox\z@\vbox{%
		\def\cr{\crcr\noalign{\kern2\p@\global\let\cr\endline}}%
		\ialign{$##$\hfil\kern2\p@\kern\@tempdima&\thinspace\hfil$##$\hfil
			&&\quad\hfil$##$\hfil\crcr
			\omit\strut\hfil\crcr\noalign{\kern-\baselineskip}%
			#1\crcr\omit\strut\cr}}%
	\setbox\tw@\vbox{\unvcopy\z@\global\setbox\@ne\lastbox}%
	\setbox\tw@\hbox{\unhbox\@ne\unskip\global\setbox\@ne\lastbox}%
	\setbox\tw@\hbox{$\kern\wd\@ne\kern-\@tempdima\left[\kern-\wd\@ne
		\global\setbox\@ne\vbox{\box\@ne\kern2\p@}%
		\vcenter{\kern-\ht\@ne\unvbox\z@\kern-\baselineskip}\,\right]$}%
	\null\;\vbox{\kern\ht\@ne\box\tw@}\endgroup}
\makeatother

\newcommand{\stepa}[1]{\overset{\rm (a)}{#1}}
\newcommand{\stepb}[1]{\overset{\rm (b)}{#1}}

\newcommand{\reals}{\mathbb{R}}

\newcommand{\Expect}{\mathbb{E}}

\newcommand{\iid}{iid\xspace}

\newcommand{\pth}[1]{\left( #1 \right)}
\newcommand{\qth}[1]{\left[ #1 \right]}
\newcommand{\sth}[1]{\left\{ #1 \right\}}

\definecolor{myblue}{rgb}{.8, .8, 1}
\definecolor{mathblue}{rgb}{0.2472, 0.24, 0.6} 
\definecolor{mathred}{rgb}{0.6, 0.24, 0.442893}
\definecolor{mathyellow}{rgb}{0.6, 0.547014, 0.24}

\newrefformat{eq}{(\ref{#1})}
\newrefformat{thm}{Theorem~\ref{#1}}
\newrefformat{th}{Theorem~\ref{#1}}
\newrefformat{chap}{Chapter~\ref{#1}}
\newrefformat{sec}{Section~\ref{#1}}
\newrefformat{seca}{Section~\ref{#1}}
\newrefformat{algo}{Algorithm~\ref{#1}}
\newrefformat{fig}{Fig.~\ref{#1}}
\newrefformat{tab}{Table~\ref{#1}}
\newrefformat{rmk}{Remark~\ref{#1}}
\newrefformat{clm}{Claim~\ref{#1}}
\newrefformat{def}{Definition~\ref{#1}}
\newrefformat{cor}{Corollary~\ref{#1}}
\newrefformat{lemma}{Lemma~\ref{#1}}
\newrefformat{prop}{Proposition~\ref{#1}}
\newrefformat{pr}{Proposition~\ref{#1}}
\newrefformat{property}{Property~\ref{#1}}
\newrefformat{app}{Appendix~\ref{#1}}
\newrefformat{apx}{Appendix~\ref{#1}}
\newrefformat{ex}{Example~\ref{#1}}
\newrefformat{exer}{Exercise~\ref{#1}}
\newrefformat{soln}{Solution~\ref{#1}}
\newrefformat{asmp}{Assumption~\ref{#1}}

\def\unifto{\mathop{{\mskip 3mu plus 2mu minus 1mu%
			\setbox0=\hbox{$\mathchar"3221$}%
			\raise.6ex\copy0\kern-\wd0%
			\lower0.5ex\hbox{$\mathchar"3221$}}\mskip 3mu plus 2mu minus 1mu}}

\ifx\lesssim\undefined
\def\simleq{{{\mskip 3mu plus 2mu minus 1mu%
			\setbox0=\hbox{$\mathchar"013C$}%
			\raise.2ex\copy0\kern-\wd0%
			\lower0.9ex\hbox{$\mathchar"0218$}}\mskip 3mu plus 2mu minus 1mu}}
\else
\def\simleq{\lesssim}
\fi

\ifx\gtrsim\undefined
\def\simgeq{{{\mskip 3mu plus 2mu minus 1mu%
			\setbox0=\hbox{$\mathchar"013E$}%
			\raise.2ex\copy0\kern-\wd0%
			\lower0.9ex\hbox{$\mathchar"0218$}}\mskip 3mu plus 2mu minus 1mu}}
\else
\def\simgeq{\gtrsim}
\fi

	\newif\ifmapx
	{\catcode`/=0 \catcode`\\=12/gdef/mkillslash\#1{#1}}
	\edef\jobnametmp{\expandafter\string\csname ic_apx\endcsname}
	\edef\jobnameapx{\expandafter\mkillslash\jobnametmp}
	\edef\jobnameexpand{\jobname}
	\ifx\jobnameexpand\jobnameapx
	\mapxtrue
	\else
	\mapxfalse
	\fi

	\renewcommand{\hat}{\widehat}
	\renewcommand{\tilde}{\widetilde}

	\theoremstyle{definition}
	
	\newtheorem{theorem}{Theorem}[section]
	\newtheorem{lemma}{Lemma}[section]

	\theoremstyle{definition}
	
	\newtheorem{assumption}{Assumption}[section]
	
	\theoremstyle{remark}
	\newtheorem{remark}{Remark}[section]
	
	\newcommand{\av}{{\boldsymbol a}}
	
	\newcommand{\xv}{{\boldsymbol x}}
	\newcommand{\Xv}{{\boldsymbol X}}
	\newcommand{\uv}{{\boldsymbol u}}
	\newcommand{\zv}{{\boldsymbol z}}
	
	\newcommand{\thetav}{\boldsymbol{\theta}}
	\newcommand{\vv}{{\boldsymbol v}}
	
	\newcommand{\Prob}{\mathbb{P}}
	
	\newcommand{\yv}{{\boldsymbol y}}
	\newcommand{\diag}{{\rm diag}}
	\newcommand{\Cc}{\mathcal{C}}
	\newcommand{\E}{\mathbb{E}}
	\newcommand{\Ec}{\mathcal{E}}
	\newcommand{\Nc}{\mathcal{N}}

	\newcommand{\wv}{{\boldsymbol w}}

	\newcommand{\xvh}{\hat{\xv}}
	\newcommand{\xvt}{\tilde{\xv}}

	\usepackage{multirow}
	
	\usepackage[textsize=tiny]{todonotes}
	
	\definecolor{darkgreen}{rgb}{0.0, 0.5, 0.0}
	
	\newcommand{\smin}{{\sigma_{\min}}}
	\newcommand{\smax}{{\sigma_{\max}}}
	\newcommand{\nm}[1]{\left\|#1\right \|}
	\newcommand{\nmhs}[1]{\left\|#1\right \|_{\sf HS}}

	\newcommand{\cb}[1]{\left\{#1\right \}}
	
	\newcommand{\emax}{{E_{\max}}}

	\newcommand{\trbr}[1]{\operatorname{Tr}\left[#1\right]}
	
\begin{document}

		
		
		\ifpdf
		\DeclareGraphicsExtensions{.pgf,.jpg,.pdf}
		\graphicspath{{figures/}{plots/}}
		\fi
		\title{Multilook Coherent Imaging: Theoretical Guarantees and Algorithms}
		
		
		\author{Xi Chen\textsuperscript{$\dagger$}, Soham Jana\textsuperscript{$\dagger$}, Christopher A. Metzler, Arian Maleki, Shirin Jalali\thanks{\textsuperscript{$\dagger$}Equal contributions. X.~C. and S.~Jalali are with the Department of Electrical and Computer Engineering, Rutgers University, New Brunswick, NJ, USA. S.~Jana is with the Department of Applied and Computational Mathematics and Statistics, University of Notre Dame, Notre Dame, IN, USA, (Correspondence to: \url{soham.jana@nd.edu}).  C.~A.~M. is with the Department of Computer Science, University of Maryland, College Park, MD, USA. A.~M. is with the Department of Statistics, Columbia University, NY, USA. 
				An earlier version of this paper was presented in part at the Proceedings of the 41st International Conference on Machine Learning, Vienna, Austria. PMLR 235, 2024.
}
		}

		\maketitle

\begin{abstract}

Multilook coherent imaging is a widely used technique in applications such as digital holography, ultrasound imaging, and synthetic aperture radar. A central challenge in these systems is the presence of multiplicative noise, commonly known as speckle, which degrades image quality. Despite the widespread use of coherent imaging systems, their theoretical foundations remain relatively underexplored. In this paper, we study both the theoretical and algorithmic aspects of likelihood-based approaches for multilook coherent imaging, providing a rigorous framework for analysis and method development.
Our theoretical contributions include establishing the first upper bound on the Mean Squared Error (MSE) of the maximum likelihood estimator under the Deep Image Prior hypothesis. Our results capture the dependence of MSE on the number of parameters in the deep image prior, the number of looks, the signal dimension, and the number of measurements per look. On the algorithmic side, we employ projected gradient descent (PGD) as an efficient method for computing the maximum likelihood solution.

Furthermore, we introduce two key ideas to enhance the practical performance of PGD. First, we incorporate the Newton-Schulz algorithm to compute matrix inverses within the PGD iterations, significantly reducing computational complexity. Second, we develop a bagging strategy to mitigate projection errors introduced during PGD updates. We demonstrate that combining these techniques with PGD yields state-of-the-art performance. Our code is available at:
 \url{{https://github.com/Computational-Imaging-RU/Bagged-DIP-Speckle}}.

\end{abstract}


\section{Introduction}\label{sec:intro}

\subsection{Problem statement}
One of the most fundamental and challenging issues faced by many coherent imaging systems is the presence of speckle noise. An imaging system with ``fully-developed" speckle noise can be modeled as
\begin{align}\label{eq:firstmodel}
\yv = A X_o \wv + \zv.
\end{align}
Here, $A\in \CC^{m\times n}$ is the sensing matrix, $\yv\in \CC^m$ is the complex-valued measurement, $X_o=\diag (\xv_o)$, where $\xv_o \in \mathbb{C}^n$ denotes the complex-valued signal, $\wv \in \mathbb{C}^n$ represents speckle (or multiplicative) noise, where $w_1,\ldots,w_n$ are independent and identically distributed (iid)  $\mathcal{CN} (\boldsymbol{0},\sigma_w^2 I_n)$, and finally  $\zv \in \CC^m$ denotes the additive noise, often caused by the sensors, and is modeled as iid $\mathcal{CN}({\boldsymbol{0}}, \sigma_z^2)$. In this paper, we explore the scenario where $m \leq n$, allowing imaging systems to capture higher-resolution images than would otherwise be possible given the number of sensors. The study of the case $m<n$ for simpler imaging systems (with no speckle noise) has led to the development of the field of compressed sensing and compressive phase retrieval \cite{donoho2006compressed, candes2008introduction, moravec2007compressive, jalali2014minimum, schniter2014compressive, bakhshizadeh2020using}. 

As is clear from \eqref{eq:firstmodel}, the multiplicative nature of the speckle noise poses a challenge in extracting accurate information from measurements, especially when the measurement matrix $A$ is ill-conditioned. To alleviate this issue, many practical systems employ a technique known as multilook or multishot acquisition \cite{argenti2013tutorial,bate2022experimental}. Instead of taking a single measurement of the image, multilook systems capture multiple measurements, aiming for each group of measurements to have independent speckle and additive noise. In an $L$-look system, the measurements captured at look $\ell$, $\ell=1,\ldots,L$, can be represented as 
\begin{align}
\label{eq:model}
\yv_\ell = A X_o \wv_\ell + \zv_\ell,
\end{align}
 where,  $\wv_1, \ldots, \wv_L \in \mathbb{C}^n$ and $\zv_1, \ldots, \zv_L \in \mathbb{C}^m$ denote  the independent  speckle noise  and additive noise vectors, respectively. In this model, we have assumed that the measurement kernel $A$ remains constant across the looks. This assumption holds true in multilook acquisition for several imaging systems, such as when the sensors' locations change slightly for different looks. 

{ Since fully-developed speckle noise is modeled as circularly symmetric complex Gaussian and has uniform phase, the phase of $\xv_o$ cannot be recovered \cite{goodman1975statistical, goodman2007speckle}. Hence, the goal of a multilook system is to obtain a precise estimate of $|\xv_o|$ based on the $L$ observations $\{\yv_1, \ldots, \yv_L\}$, given the measurement matrix $A$. (Here, $|\cdot|$  denotes the element-wise absolute value operation.) Therefore, we assume that $\xv_o$ is real-valued.
More specifically, we consider the recovery of the real-valued speckle-free reflectivity $\xv_o \in \mathbb{R}^n$ from complex-valued measurements $\yv \in \mathbb{C}^m$ acquired by a coherent imaging system \cite{pellizzari2018optically, pellizzari2020coherent}. Here, $\xv_o^2$ corresponds to the intensity image observed under incoherent illumination, i.e., the type of image commonly perceived in everyday vision. This is the primary quantity of interest to recover in applications such as digital holography and synthetic aperture radar \cite{argenti2013tutorial, bianco2018strategies}.}

A standard approach for estimating $\xv_o$ is to minimize the negative log-likelihood function subject to the signal structure constraint. More precisely, in a constrained likelihood-based approach, one aims to solve the following optimization problem:
\begin{equation}\label{eq:formulation1}
\xvh = \arg\min_{\xv \in \Cc} f_L(\xv),
\end{equation}
 where $\Cc$ represents the set encompassing all conceivable images and $f_{L}(\xv)$ is defined as:
\begin{align}
    f_{L}(\xv) 
    =&  \log \det(B(\xv)) + \frac{1}{L} \sum_{\ell=1}^L \tilde{\yv}_\ell^{\top} (B(\xv))^{-1} \tilde{\yv}_\ell, \label{eq:complex-likelihood}
\end{align}
where  
\begin{align}\label{eq:complex-cov}
    B(\xv) = \begin{bmatrix} \sigma_z^2 I_n + {\sigma_w^2} \Re (U(\xv)) & -{\sigma_w^2} \Im (U(\xv)) \\ {\sigma_w^2} \Im (U(\xv)) & \sigma_z^2 I_n + {\sigma_w^2} \Re (U(\xv)) \end{bmatrix},
\end{align}
and
$\tilde{\yv}_\ell^{\top}  = \begin{bmatrix}
        \Re (\yv_\ell^{\top}) & \Im (\yv_\ell^{\top})
    \end{bmatrix},
X=\diag(\xv)$ and $U(\xv)=A X^2 \bar{A}^{\top}$.
Here, $\Re(\cdot)$ and $\Im(\cdot)$ denote element-wise real and imaginary parts, respectively. (Appendix \ref{app:MLE} presents the derivation of the log-likelihood function and its gradient.)

It is important to note that the set $\Cc$ in \eqref{eq:formulation1} is not known explicitly  in practice. Hence, in this paper, we work with the following hypothesis that was put forward in \cite{ulyanov2018deep, heckel2018deep}:  Natural images can be embedded within the range of untrained neural networks that have substantially fewer parameters than the total number of pixels, and use \iid noise as input. 
We call this hypothesis the Deep Image Prior (DIP) Hypothesis and formalize it in Section \ref{sec:main-theory}.  Using the DIP hypothesis, this paper focuses on addressing two main challenges:

\begin{itemize}
\item Theoretical challenge: Assuming that we can solve the optimization problem \eqref{eq:formulation1} under the DIP hypothesis, the following question arises: Can we theoretically characterize the corresponding reconstruction quality? Moreover, what is the relationship between the reconstruction error and key parameters such as $k$ (the number of parameters in the DIP neural networks), $m$, $n$ and $ L$?

\item Practical challenge: Given the challenging nature of the likelihood and the DIP hypothesis, can we design a computationally-efficient algorithm for solving \eqref{eq:formulation1}?
\end{itemize}

\subsection{Contributions}
We summarize our contributions as follows:

\subsubsection{Theoretical contribution} We establish the first theoretical result on the performance of multilook coherent imaging systems. Compared to the conference version of this paper, namely \cite{chenbagged}, the theoretical results presented here are significantly sharper.
Our findings unveil intriguing characteristics of such imaging systems. A special case of our result, corresponding to \( L = 1 \), is directly comparable to the findings in \cite{zhou2022compressed}. As we will show, in this setting, our bounds on the mean squared error are significantly tighter than those presented in \cite{zhou2022compressed}.

\subsubsection{Algorithmic contribution} On the practical side, we start with vanilla projected gradient descent (PGD) \cite{lawson1995solving}, which faces two challenges that diminish its effectiveness on this problem:

\begin{itemize}

\item Challenge 1 and its solution: As will be described in Section \ref{sec:PGDchallenges}, in the PGD, the signal to be projected onto the range of the untrained neural network is buried in ``noise''. Hence, DIP with a large number of parameters will overfit to the noise and will not allow the PGD algorithm to obtain a reliable  estimate \cite{heckel2018deep,heckel2019denoising}. On the other hand, the low accuracy of simpler DIP becomes a bottleneck as the algorithm progresses through iterations, limiting the overall performance. To alleviate this issue, we propose \textbf{Bagged-DIP}. This is a simple idea with roots in classical literature of ensemble methods \cite{breiman1996bagging}. The Bagged-DIP idea enables us to use complex DIP at every iteration and yet obtain accurate results.
\item Challenge 2 and its solution: As will be clarified in Section \ref{ssec:pgd:dip:intro}, PGD requires the inversion of large matrices at every iteration, which is a computationally challenging problem. We alleviate this issue by using the Newton-Schulz algorithm \cite{schulz1933iterative}, and empirically demonstrating that \textbf{only one} step of this algorithm is sufficient for the PGD algorithm. This significantly reduces the computational complexity of each iteration of PGD. 

\end{itemize}
{
\subsection{Notations}
For a matrix $A$, let $\smax(A)$ and $\smin(A)$ denote the maximum and minimum singular values of $A$ and, if $A$ is symmetric, $\lambda_1(A)\leq\dots\leq \lambda_n(A)$ denote the eigenvalues of $A$. Boldface lowercase letters such as $\xv$ are used for vectors. Furthermore $\|\cdot\|_2$ denotes $\ell_2$ norm or the operator norm depending on the argument being a vector or matrix, $\|\xv\|_\infty$ denotes the $\ell_\infty$ norm of the vector $\xv$, and $\nmhs{A}$ denotes the Hilbert-Schmidt norm of a matrix $A$. For a vector $\xv=(x_1,\dots,x_n)$, we let $\xv^2:=(x_1^2,\dots,x_n^2).$ Given two sequences $\{a_n\}_{n=1}^\infty$ and $\{b_n\}_{n=1}^\infty$, we say $a_n=O(b_n)$, or equivalently $b_n=\Omega(a_n)$, if there exist constants $C>0$, and $M>0$, such that for all $n>M$, $|a_n|\le C |b_n|$. We say $a_n=o(b_n)$, or equivalently $b_n=\omega(a_n)$, if $\lim_{n\to\infty} {a_n}/{b_n} =0$. We write $a_n = \Theta (b_n)$ if $a_n = O(b_n)$ and $b_n = O(a_n)$. Throughout the manuscript, $\calC\calN(\cdot,\cdot)$ will denote the complex Gaussian distribution and $\calN(\cdot,\cdot)$ will denote the Gaussian distribution over real space. Define the unit balls in $\reals^k$ with respect to the supremum norm as $B_{\infty,k} \left(\mathbf{0}, r\right) = \{ \thetav \in \mathbb{R}^k \ | \  \|\thetav\|_\infty \leq r\}$.

\subsection{Organization of the paper}
We review the related work and compare our results with the existing literature in \prettyref{sec:related-work}. In \prettyref{sec:main-theory}, we introduce our measurement model and assumptions for theoretical analysis, and present our main theorem. \prettyref{sec:mainempirical} presents the algorithmic contributions including the efficient gradient calculation and the Bagged-DIP-based PGD framework. In \prettyref{sec:simulations}, we provide a collection of numerical studies to support the effectiveness of our proposed algorithm. The proof of our main theorem is provided in \prettyref{sec:proof-theory}. The auxiliary technical results and more detailed computational components are presented in the appendix.}

\section{Related Work}
\label{sec:related-work}
Eliminating speckle noise has been extensively explored in the literature~\cite{lim1980techniques,gagnon1997speckle,tounsi2019speckle}. Current technology relies on gathering enough measurements to ensure the invertibility of the sensing matrix $A$ and subsequently inverting $A$ to represent the measurements in the following form: $\yv_\ell = X\wv_\ell+ \zv_\ell$.
However, as the matrix $A$ deviates from the identity, the elements of the vector $\zv$ become dependent. In practice, these dependencies are often overlooked, simplifying the likelihood function of the measurements. This simplification allows researchers to leverage various denoising methods, spanning from classical low-pass filtering \cite{lee1980digital, kuan1985adaptive} to the application of convolutional neural networks \cite{tian2020attention} and transformers \cite{fan2022sunet}. A series of papers have considered the impact of the sensing matrix in the algorithms. By using single-shot digital holography, the authors in~\cite{pellizzari2017phase, pellizzari2018optically} developed a heuristic method to obtain a maximum a posteriori estimate of the real-valued speckle-free object reflectance. They later extend this method to handle multi-shot measurements and incorporate more accurate image priors~\cite{pellizzari2020coherent,pellizzari2022solving, bate2022experimental}. While these methods can work with non-identity $A$'s, they still require or assume $A$ to be well-conditioned.

Our paper is different from the existing literature, mainly because we study scenarios where the matrix $A$ is under-sampled ($m<n$). In a few recent papers, researchers have explored similar problems~\cite{zhou2022compressed,chen2023multilook}. \cite{chen2023multilook} is closely related to this work in both scope and methodology, addressing a similar problem while advocating the use of DIP-based PGD. Building on this prior effort, the proposed Bagged-DIP-based PGD with the Newton–Schulz algorithm achieves substantial improvements in both reconstruction quality and computational efficiency compared to \cite{chen2023multilook}. Furthermore, while \cite{chen2023multilook} serves as an initial exploration of this direction, it does not provide theoretical guarantees for the performance of DIP-based MLE. In contrast, the present work develops a more rigorous and comprehensive theoretical analysis that formalizes and extends those earlier insights.

 The authors  in \cite{zhou2022compressed} theoretically demonstrated the feasibility of accurate recovery of ${\bx}_o$ even for  $m<n$ measurements. While our theoretical results build upon the contributions of \cite{zhou2022compressed}, our paper extends significantly in three key aspects:  (1) We address the multilook problem and investigate the influence of the number of looks on our bounds. To ensure sharp bounds, especially when $L$ is large, we derive sharper bounds than those presented in \cite{zhou2022compressed}. These require a different proof technique as detailed in our proof. (2) In contrast to the use of compression codes' codewords for the set $\Cc$ in \cite{zhou2022compressed}, we leverage the range of a DIP, inspired by recent advances in machine learning. Despite presenting new challenges in proving our results, this approach enables us to simplify and establish the relationship between Mean Squared Error (MSE) and problem specification parameters such as $n, m, k, L$.
(3) On the empirical side, the experiments in \cite{zhou2022compressed} were restricted to a few toy examples, such as a piece-wise constant signal, due to their limiting assumptions. In contrast, by leveraging DIP together with the Newton–Schulz method and bagging, we are able to evaluate our algorithms on natural images and achieve the state-of-the-art results.

Given DIP's flexibility, it has been employed for various imaging inverse problems, e.g., compressed sensing, phase retrieval, etc.~\cite{jagatap2019algorithmic, ongie2020deep,darestani2021accelerated, ravula2022one,zhuang2022practical,zhuang2023blind}. 
To boost the performance of DIP in these applications, researchers have explored several ideas, including, introducing explicit regularization \cite{mataev2019deepred}, incorporating prior on network weights by introducing a learned regularization method into the DIP structure \cite{van2018compressed}, combining with pre-trained denoisers in a Plug-and-Play fashion \cite{zhang2021plug, sun2021plug}, and exploring the effect of changing DIP structures and input noise settings to speed up DIP training \cite{li2023deep}.

Lastly, it's important to note our work can be situated within the realm of compressed sensing (CS) \cite{donoho2006compressed, candes2008introduction, davenport2012introduction, bora2017compressed, peng2020solving, joshi2021plugin, nguyen2022provable}, where the objective is to derive high-resolution images from lower-resolution measurements. However, notably, the specific challenge of recovery in the presence of speckle noise has not been explored in the literature before, except in \cite{zhou2022compressed} that we discussed before.


\section{Main Theoretical Result}\label{sec:main-theory}

\subsection{Model, assumptions and their justifications}

{ We now present the theoretical analysis of estimating $\bx_o \in \mathbb{R}^n$ from the measurements acquired from the measurement model \eqref{eq:model}. To simplify the theoretical analysis, we analyze the case in which all the complex-valued components ($A,\wv_\ell$, thus $\yv_\ell$) are replaced by the corresponding real-valued counterparts, and assume no additive noise exists, i.e., $\sigma_z=0$. The choice of $\sigma_z=0$ is motivated by the observation that, in coherent imaging settings, additive noise is often negligible compared to speckle noise. For instance, \cite{tippie2012weak} notes that heterodyne detection in digital holography suppresses auxiliary noise sources, rendering them insignificant relative to the signal-induced noise. Accordingly, speckle noise is recognized as the dominant noise component in many imaging applications \cite{lucke2002photon}. 
On a higher level, we expect that the theoretical analysis is similar for the complex-valued components with general $\sigma_z$. 
Our algorithm for the numerical analysis in \prettyref{sec:simulations} is presented to optimize the complex-valued likelihood in \eqref{eq:complex-likelihood} when additive noise is present. As the simulations in \prettyref{sec:simulations} will also confirm, the impact of additive noise on the results is relatively mild, while speckle noise has the dominant effect. }

{ 
	In summary, in this section, we work with the real-valued likelihood function corresponding to the model
\begin{align}\label{eq:model_theory}
	\yv_\ell = A X_o \wv_\ell,
	\quad \ell=1,\dots,L,
\end{align}
where $	\yv_{\ell} \in \reals^m$, $A\in \reals^{m\times n}$, and $\bw_\ell\sim \calN(0,\sigma_w^2 I_n)$.
Given $A$, the maximum likelihood estimator over a real-valued functional class $\calC$ (here corresponds to the set of images of interest) is 
\begin{equation}\label{eq:updated_lik}
\xvh = \argmin_{\xv \in \Cc} f(\xv),
\end{equation}
where 
\begin{equation}
f(\xv)=  \log\det \left(\sigma_w^2 AX^2A^{\top}\right)  
+ \frac{1}{L} \sum^L_{\ell=1} \yv_{\ell}^{\top} \left(\sigma_w^2 AX^2A^{\top}\right)^{-1}  \yv_{\ell}.
\end{equation}
Note that we omit subscript $L$ from the likelihood as a way to distinguish the negative log-likelihood of real-valued measurements from the complex-valued ones. Before stating our theoretical results, we would like to formally state the DIP hypothesis that was heuristically described in Section \ref{sec:intro}. 

\begin{assumption}[DIP hypothesis based characterization of image class with effective dimension $k$]\label{asmp:diphyp}
Let $\uv \in \mathbb{R}^d$ denote a vector with iid $\calN(0,1)$ elements. With high probability on the choice of $\uv$, there exists a function $g_{\thetav} (\uv)$ parameterized by $\thetav \in \mathbb{R}^k$ that satisfies the following conditions:
\begin{enumerate}
\item[(a)] As a function of {$\thetav$}, $g_{\thetav} (\uv)$ is Lipschitz with Lipschitz constant $1$.

\item[(b)] $g_{\mathbf{0}}(\uv) = \mathbf{0}$.

\item[(c)] The set of images of interest $\calC$ satisfies:
\begin{align}\label{eq:dip:hyp}
        \calC \subset 
        \{
        &\xv \in \reals^n\ | \ \xv= g_{\thetav} (\uv), \thetav \in B_{\infty,k}\pth{\mathbf{0}, x_{\max}\sqrt{\frac nk}},\notag \\ 
        &0 < x_{\min} \leq x_{i} \leq x_{\max},1\leq i\leq n\},
 \end{align}
 where $x_{\min}$ and $x_{\max}$ are two constants. 
\end{enumerate}

\end{assumption}

{
In the following remarks, we would like to clarify some aspects of \prettyref{asmp:diphyp}. 

\begin{remark}
  As discussed earlier, in practice, $g_{\thetav}(\cdot)$ is typically constructed by a neural network with parameters $\thetav$. The input to this network, namely the vector $\uv$, is preselected as iid draws from $\cN(0,1)$. However, many other choices of distributions, such as uniform or Laplace distributions, can also be used effectively. Furthermore, we should also emphasize that the condition \( g_{\mathbf{0}}(\uv) = \mathbf{0} \), holds for all networks typically used in the DIP framework. 
\end{remark}

\begin{remark}

The neural networks used to map random noise to images are typically not based on fully connected layers alone. Instead, they often rely on convolutional layers, upsampling operators, and other architectural components that introduce strong inductive biases such as locality while using substantially fewer parameters. For example,  Figure~\ref{fig:DIP_block_structure} of Appendix~\ref{app:add_experiment} illustrates the basic Deep Decoder structure used in our simulations.
Prior work on architectures such as the Deep Decoder \cite{heckel2018deep,heckel2019denoising} suggests that, in many realistic noisy settings, reducing parameter count through architectural constraints or fewer channels can still maintain strong performance, although the precise effect depends on the task and reconstruction setting.
     More specifically, as noted in \cite{heckel2018deep} (p.~5), effective reconstruction of an RGB image with $786{,}432$ pixels can be achieved using the Deep Decoder with a constant number of channels across convolutional layers. When the number of channels is set to 64 and 128, the corresponding parameter counts are $25{,}536$ and $100{,}224$, respectively, both significantly smaller than the number of image pixels. This observation suggests that even when $k$ is much smaller than $n$ (where $k$ is the number of network parameters and $n$ the number of pixels), the network can still represent natural images effectively.
	
\end{remark}

\begin{remark}
One unconventional aspect of \eqref{eq:dip:hyp} is that, we allow the size of $\theta_i$ to grow with $n$. This is closely related to our assumption that $g_{\thetav}(\uv)$ is a Lipschitz function of $\thetav$ with Lipschitz constant $1$. To clarify this point, using the Lipschitzness and the fact that $g_{\mathbf{0}}(\uv) = \mathbf{0} $, we have
\begin{equation}\label{eq:size_theta_1}
\|g_{\thetav}(\uv)\|_2 \leq \|\thetav\|_2 \leq \sqrt k \|\thetav\|_{\infty}. 
\end{equation}
On the other hand, we assume that
$0 < x_{\min} \leq x_{0,i} \leq x_{\max}$ to study the detection in the setup of bounded and non-zero signals. In particular, the need for a strictly positive $x_{\min}$ is purely due to technical reasons related to our proofs. Therefore,
\begin{equation}\label{eq:size_theta_2}
x_{\min}\sqrt{n}
\leq
\|\xv\|_2
=
\|g_{\thetav}(\uv)\|_2
\leq
x_{\max}\sqrt{n}.
\end{equation}
Combining \eqref{eq:size_theta_1} and \eqref{eq:size_theta_2}, we have
\[
\sqrt{k}\|\thetav\|_{\infty}
\geq
\|\thetav\|_2
\geq
x_{\min}\sqrt{n}.
\]
This means that $\|\thetav\|_\infty = \Omega (\sqrt{\frac{n}{k}})$. Note that while this constraint suggests that to keep $g_{\thetav} (\cdot)$ Lipschitz with Lipschitz constant one, $\|\thetav\|_{\infty}$ should grow with $n$, it does not necessarily impose any upper bound on $\|\thetav\|_\infty$. However, to keep the notation simpler and since we do not want to make the elements of $\thetav$ unnecessarily large, we have also assumed that $\|\thetav\|_{\infty} \leq x_{\max} \sqrt{\frac{n}{k}}$. However, we should emphasize that the impact of this upper bound on our theoretical results is minor, and one can consider larger radii (as long as they remain polynomials of $n/k$) without major impact on our theoretical results.  
\end{remark}
}

The following assumptions summarize all the conditions required for the theoretical results

\begin{assumption}[Measurement model]\label{asmp:sensor-signal}
	The known $m \times n$ sensing matrix $A$ and the measurements satisfy the following.
	\begin{enumerate}[label=(\roman*)]
		\item The sensing matrix is under-determined, i.e., $m<{n\over 1+ \xi_0}$ for some constant $\xi_0>0$.
		\item The entries of $A$ are generated according to $A_{ij}\simiid \calN(0,1),i=1,\dots, m, j=1,\dots,n$. 
        \item The measurements obey model~\eqref{eq:model_theory}, i.e., there is no additive noise component. 
	\end{enumerate}
\end{assumption}

}

\subsection{Main theorem and its implications}
We are now ready to state our main theorem. Our main theorem captures the interplay between the accuracy of our maximum likelihood-based recovery, the number of measurements $m$, the number of looks $L$, the ambient dimension of the signals to be recovered $n$, and the number of the parameters in the DIP model. 

\begin{theorem}\label{thm:maintheorem}
	Suppose that $m$, $n$ and $L$ denote the number of measurements, the ambient signal dimension, and the number of looks $L$, respectively. Suppose that the true signal $\xv_o$ is embedded in a space of dimension $k$, as given in \prettyref{asmp:diphyp}, and the sensing matrix $A$, and the measurements satisfy \prettyref{asmp:sensor-signal}. Then, the maximum likelihood estimator $\xvh$ of $\xv_o$ given in \eqref{eq:updated_lik} satisfies
	\begin{align}\label{eq:thm:MSEbound}
		\frac{1}{n} \|\xvh-\xv_o\|_2^2 \leq   C_1\pth{{n\over m^2}\cdot {k\log n\over L}+{\sqrt{k \log{n}}\over m}},
	\end{align}
	with probability 
	$$1- C_2\pth{e^{-C_3m} + e^{-C_3Ln} + e^{-C_3k \log n}+   e^{C_4k \log n-C_3n}}$$ for constants $C_1,C_2,C_3,C_4>0$. 
\end{theorem}

Before discussing the proof sketch and the technical novelties of our proof strategy, we explain some of the conclusions that can be drawn from this theorem, provide some intuition, and compare with some of the existing results. Our first remark compares our result with the only existing theoretical work on this problem:

\begin{remark}
	A much weaker version of Theorem \ref{thm:maintheorem} appeared in \cite{chenbagged}. Specifically, the leading term in the upper bound was $\frac{n}{m} \cdot \frac{\sqrt{k \log n}}{\sqrt{L m}}$, which is substantially looser than the bound obtained here, particularly since $k \log n$ is much smaller than $Lm$. The sharper result presented in this paper is derived using a different proof strategy.
\end{remark}

The only existing theoretical results on the recovery performance of coherent imaging systems are those presented in \cite{zhou2022compressed,malekian2025speckle}. The authors of \cite{malekian2025speckle} focus exclusively on the despeckling problem and do not address the central challenge considered in our work -- namely, the presence of the measurement matrix \( A \) and the fact that \(m < n \). As a result, their results are not directly comparable to ours. However, the results of \cite{zhou2022compressed} are more closely related to our work. But there are a few major differences between the theoretical result presented in \cite{zhou2022compressed} and Theorem \ref{thm:maintheorem}:
\begin{enumerate}
	\item The theoretical results in \cite{zhou2022compressed} rely on the existence of a compression algorithm tailored to the set of images, whereas our work is based on the DIP hypothesis. The DIP hypothesis is more flexible and, as demonstrated in the simulation section, allows for efficient solutions to the associated optimization problem.
	
	\item While there are some major differences, we interpret the \( \alpha \)-dimension of the sequence of compression algorithms as the effective number of parameters $k$ so that we can provide a more accurate comparison between our result and the one presented in \cite{zhou2024correction,zhou2022compressed}. The authors in the above works consider the setup where $m,n$ are of the same orders, $L=1$, and their upper bound \cite[Corollary 1]{zhou2024correction} takes the form \( \sqrt{\frac{k \log n}{m}}\). Comparing this with our bound in the special case \( L = 1 \), we observe that when $\frac{k \log n}{m}$ is much smaller than 1, our bounds are significantly sharper than those presented in \cite{zhou2022compressed}. In fact, it is straightforward to see that since $n>m$ and $n$ is proportional to $m$, our upper bound in Theorem \ref{thm:maintheorem} simplifies to  \( \frac{k \log n}{m} \). One source of looseness in the proof of \cite{zhou2022compressed} arises from an early step that involves working with the expected log-likelihood. To overcome this issue, we develop a proof strategy that entirely avoids relying on the expected log-likelihood.
	
	\item This paper also establishes the dependence of the error on the number of looks.

\end{enumerate}

Now, we provide more information on the upper bound we have obtained in Theorem \ref{thm:maintheorem}. As is clear in \eqref{eq:thm:MSEbound}, there are two terms in the MSE. One that does not change with $L$ and the other term that decreases with $L$. To understand these two terms, we provide further explanation in the following remarks.

\begin{remark}
	As the number of parameters of DIP, $k$, increases (while keeping $m, n$, and $L$ fixed), both error terms in the upper bound of MSE grow. This aligns with intuition, as increasing the number of parameters in $g_{\thetav}(\uv)$ allows the DIP model to generate more intricate images. Consequently, distinguishing between these diverse alternatives based on the noisy measurements becomes more challenging.
\end{remark}

\begin{remark}
	The main interesting feature of the first term in the MSE, i.e., $\frac{n  k \log n}{  Lm^2}$,  is the fact that it grows rapidly as a function of $n$. In imaging systems with only additive noise, the growth is often logarithmic in $n$ \cite{bickel2009simultaneous}, contrasting with polynomial growth observed here. This can be attributed to the fact that as we increase $n$, the number of speckle noise elements present in our measurements also increases. Hence, it is reasonable to expect the error term to grow faster in $n$ compared with additive noise models. However, the exact rate at which the error increases is yet unclear. Nonetheless, we believe the dependency on $L$ for the term $\frac{n  k \log n}{   L m^2}$ is sharp as it aligns with the notion of parametric error rate $\frac 1L$ for an estimation problem with $L$ samples.   
\end{remark}

\begin{remark}
	As $L \rightarrow \infty$, the first term in the upper bound of MSE converges to zero, and the dominant term becomes $\sqrt{k \log n}/m$. Note that since we are considering a fixed matrix $A$ across the looks, even when $L$ goes to infinity, we should not expect to be able to recover $\xv_o$ independent of the value of $m$. One heuristic way to see this is to calculate 
	\begin{align*}
		\frac{1}{L} \sum_{\ell=1}^L \yv_{\ell} \yv_{\ell}^{\top} =  AX_o \frac{1}{L}\sum_{\ell=1}^L \wv_{\ell} \wv_{\ell}^{\top} X_oA^{\top}.
	\end{align*} 
	If we heuristically apply the weak law of large numbers and approximate $\frac{1}{L}\sum_{\ell} \wv_{\ell} \wv_{\ell}^{\top}$ with $I$, we get
	\[
	\frac{1}{L} \sum^L_{\ell=1} \yv_{\ell} \yv_{\ell}^{\top} \approx AX_o^2 A^{\top}.
	\]
	Under these approximations, the matrix $\frac{1}{L} \sum_{\ell} \yv_{\ell} \yv_{\ell}^{\top}$ provides $m(m+1)/2$ (due to symmetry) linear measurements of $X_o^2$. Hence, inspired by classic results in compressed sensing \cite{candes2013well}, intuitively, we expect the accurate recovery of $\xv_o^2$ to be possible when $m^2$ is much larger than $k \log n$. The first error term in MSE is negligible when $m^2$ is much larger than $k \log n$, which is consistent with our conclusion based on the limit of $\frac{1}{L} \sum_{\ell} \yv_{\ell} \yv_{\ell}^{\top}$. 
\end{remark}

We next provide the key steps of the proof to highlight the technical novelties of our proof and also to enable the readers to navigate through the detailed proof more easily. We follow it up with the details to end this section.

\subsection{Key steps in the proof of Theorem \ref{thm:maintheorem}} \label{ssec:proof:sketch}

We will use the following notations throughout the proof of our main theorem:
\begin{align}
	\label{eq:proof-notations}
	\begin{gathered}
		\xv_o\in \reals^n, \;X_o=\diag(\xv_o),\; \Sigma_o=(A X_o^2A^{\top})^{-1},\\
		\xvh\in \reals^n,\; \hat X=\diag(\hat \bx),\; \hat\Sigma=(A\hat X^2A^{\top})^{-1},
		\; \Delta \Sigma = \hat \Sigma-\Sigma_o,\\
		\xvt\in \reals^n,\; \tilde X=\diag(\tilde \bx),\; \tilde\Sigma=(A\tilde X^2A^{\top})^{-1},
		\; \Delta\tilde \Sigma = \tilde \Sigma-\Sigma_o,\\
		\xv\in \reals^n,\; X=\diag(\xv), \; \Sigma=(AX^2A^\top)^{-1}.
	\end{gathered}
\end{align}
 Denote the objective function $f$ as
\begin{align}\label{eq:likelihood}
	f(\xv)=f(\Sigma)=-\log\det \Sigma +{1\over L \sigma_w^2}\sum_{\ell =1}^L{\rm Tr}(\Sigma\yv_{\ell}\yv_{\ell}^{\top})
.
\end{align} 
For a given $\Sigma$, let $\bar{f}(\Sigma)$ denote the expected value of  ${f}(\Sigma)$ with respect to $\sth{\wv_{\ell}}_{\ell=1}^L$. It is straightforward to show
\begin{align}\label{eq:min-f-bar}
	\bar{f}(\Sigma)=-\log\det \Sigma +{\rm Tr}(\Sigma AX_o^2A^{\top}).
\end{align}
Let $\xv_o$ denote the true signal and $\hat{\xv}$ denote the minimizer of the objective $f$. Then, we have
\begin{align}\label{eq:thm1-step1}
	f(\hat{\Sigma})\leq f({\Sigma}_o),
\end{align}
Expanding the terms in \eqref{eq:thm1-step1} we have
\begin{align}
	\label{eq:m12}
	&\frac 1{ L\sigma_w^2} \sum_{\ell=1}^L\yv_\ell^{\top} \hat{\Sigma}\yv_\ell-\log \det(\hat {\Sigma})
	\leq 
	\frac 1{ L\sigma_w^2} \sum_{\ell=1}^L\yv_\ell^{\top} {\Sigma}_o\yv_\ell-\log \det({\Sigma}_o).
\end{align}
Using the above definition and \eqref{eq:min-f-bar}, we reorganize the terms in \eqref{eq:m12} to obtain
\begin{align}\label{eq:key:equation}
	\bar f(\hat{\Sigma})-\bar f({\Sigma}_o)
	\leq -\qth{\frac 1{L\sigma_w^2} \sum_{\ell=1}^L \yv_\ell^{\top} \Delta{\Sigma}\yv_\ell-\Tr(\Delta{\Sigma}{\Sigma}_o^{-1})}.
\end{align}

The rest of the proof can be summarized in the following steps:

\begin{enumerate}
\item We establish a lower bound for $\bar f(\hat{\Sigma})-\bar f({\Sigma}_o)$ in terms of $\Tr({\Sigma}_o^{- 1}\Delta{\Sigma}{\Sigma}_o^{-1}\Delta{\Sigma})$. 

\item We obtain an upper bound for $-\qth{\frac 1{L\sigma_w^2} \sum_{\ell=1}^L \yv_\ell^{\top} \Delta{\Sigma}\yv_\ell-\Tr(\Delta{\Sigma}{\Sigma}_o^{-1})}$ in terms of $\Tr({\Sigma}_o^{- 1}\Delta{\Sigma}{\Sigma}_o^{-1}\Delta{\Sigma})$. 

\item We use the two bounds derived in Steps 1 and 2 to obtain an upper bound for $\Tr({\Sigma}_o^{- 1}\Delta{\Sigma}{\Sigma}_o^{-1}\Delta{\Sigma})$.

\item We use concentration arguments to obtain a high probability lower bound (based on the randomness in $A$) for $\Tr({\Sigma}_o^{- 1}\Delta{\Sigma}{\Sigma}_o^{-1}\Delta{\Sigma})$ in terms of $\|\xvh-\xv_o\|_2^2$. By combining it with Step 3, we establish a high-probability error upper bound on $\|\xvh-\xv_o\|_2^2$.  
\end{enumerate}
The main difficulty in the analysis arises from the fact that the quantities \(\hat{\xv}\), \(\hat{\Sigma}\), and \(\Delta\Sigma\), which emerge from the optimization problem in \eqref{eq:thm1-step1}, depend on the observations and underlying random variables in a highly coupled and nonlinear manner. Because of these intricate dependencies, it is difficult to directly apply standard probabilistic and statistical tools.

To overcome this challenge, we first employ decoupling arguments to simplify the dependence structure of the relevant random quantities. We then construct suitable covering sets for the signal class, which enables us to control the estimation error uniformly over the parameter space. Additional details regarding these steps are provided in \prettyref{sec:proof-theory}.

\section{Main Algorithmic Contributions}\label{sec:mainempirical}
\subsection{Summary of DIP-based Projected Gradient Descent}\label{ssec:pgd:dip:intro}

Note that the algorithm and the corresponding results are developed for the problem formulated in \eqref{eq:formulation1}, which is in the complex-valued setting. As discussed in Section \ref{sec:intro}, we aim to solve the constrained optimization problem \eqref{eq:formulation1} under the DIP hypothesis. A popular heuristic for achieving this goal is to use projected gradient descent (PGD). At each iteration $t$, the estimate ${\bx}^t$ is updated as follows:
\begin{equation}
    {\xv}^{t+1} = \text{Proj} \left(\xv^t - \mu_t \nabla f_L({\xv}^t) \right),
\end{equation}
where $\text{Proj} (\cdot)$ projects its input onto the range of the function $g_{\thetav} (\uv)$, and $\mu_t$ denotes the step size. The details of the calculation of $\nabla f_L({\bx}^t)$ are outlined in Appendix \ref{app:MLE}.

An outstanding question in the implementation pertains to the nature of the projection operation $\text{Proj} (\cdot)$. If $g_{\thetav}(\uv)$, in which $\thetav$ denotes the parameters of the neural network and $\uv$ denotes the input Gaussian noise, represents the reconstruction of the DIP, during training, DIP learns to reconstruct images by performing the following two steps:
\begin{align} \label{eq:PGD}
&\hat{\thetav}^t = \operatorname*{argmin}_{\thetav} \left\| g_{\thetav}(\uv) - (\xv^t - \mu_t \nabla f_L({\xv}^t)) \right\|^2_2, \nonumber \\
&{\xv}^{t+1} = g_{\hat{\thetav}^t}(\uv), 
\end{align}
where to obtain a local minima in the first optimization problem, we use Adam optimizer \cite{kingma2014adam} for learning the parameters $\thetav$.  
One of the main challenges in using DIP in PGD is that the performance of DIP $g_{\thetav}(\uv)$ is affected by the structure choices, training iterations as well as the statistical properties of $\xv^t - \mu_t \nabla f_L({\xv}^t)$~\cite{heckel2019denoising}. We will discuss this issue in the next section.

\subsection{Challenges in DIP-based PGD and the Solutions} \label{sec:PGDchallenges}

In this section, we examine the two primary challenges encountered in applying the DIP-based PGD, and present novel perspectives for addressing them.

\subsubsection{Challenge 1: Right choice of DIP}

Designing PGD, as described in Section \ref{ssec:pgd:dip:intro}, is particularly challenging when it comes to selecting the appropriate network structure for DIP. Figure \ref{fig:decoder_compare_clean} in Appendix~\ref{app:DIP} clarifies the main reason. In this figure, four DIP networks are used for fitting to the clean image (left panel) and an image corrupted by the Gaussian noise (right panel). As is clear, the sophisticated networks fit the clean image very well. However, they are more susceptible to overfitting when the image is corrupted with noise. On the other hand, the networks with simpler structures do not fit the clean image well but are less susceptible to noise than the sophisticated DIP. This issue has been observed in previous work \cite{heckel2018deep, heckel2019denoising}. 

The problem outlined above poses a challenge for the DIP-based PGD. Note that if ${\bx}^t - \mu_t \nabla f_L({\bx}^t)$ closely approximates ${\bx}_o$, fitting a highly intricate DIP to ${\bx}^t - \mu_t \nabla f_L({\bx}^t)$ will yield an estimate that remains close to ${\bx}_o$. Conversely, if overly simplistic networks are employed in this scenario, their final estimate may fail to closely approach ${\bx}^t - \mu_t \nabla f_L({\bx}^t)$, resulting in a low-quality estimate. In the converse scenario, where ${\bx}^t - \mu_t \nabla f_L({\bx}^t)$ is significantly distant from ${\bx}_o$, a complex network may overfit to the noise part. On the contrary, a simpler network, capable of learning only fundamental features of the image, may generate an estimate that incorporates essential image features, bringing it closer to the true image.

The above argument suggests the following approach: initiate the DIP used in PGD with simpler networks and progressively shift towards more complex structures as the estimate quality improves\footnote{A somewhat weaker approach would be to use intricate networks at every iteration, but then use some regularization approach such as early stopping to control the complexity of the estimates. }. However, finding the right complexity level of the DIP for each iteration of PGD, in which the statistics of the error in the estimate ${\bx}^t - \mu_t \nabla f_L({\bx}^t)$ is not known and may be image dependent, is a challenging problem. Next, we propose a novel approach to addressing this issue. 

\subsubsection{Solution to Challenge 1: Bagged-DIP}\label{ssec:baggedDIP}

Our new approach is based on a classical idea in statistics and machine learning: Bagging. 
Rather than finding perfectly the right complexity level for the DIP at each iteration, which is a computationally demanding and statistically challenging problem, we apply bagging to DIP, which we term it as Bagged-DIP. The idea of bagging is that in the case of challenging estimation problems, we create several low-bias and, hopefully, weakly dependent estimates  (we are overloading the phrase weakly dependent to refer to situations in which the cross-correlations of different estimates are not close to $1$ or $-1$)  of a quantity and then calculate the average of those quantities to obtain a lower-variance estimate. In order to obtain weakly dependent estimates, a common practice in the literature is to apply the same learning scheme to multiple datasets, each of which is a random perturbation of the original training set, e.g., the construction of random forests. 

{ While there are many ways to construct Bagged-DIP estimates for projecting $\xv^t - \mu_t \nabla f_L(\xv^t)$, we focus here on a simple strategy to create a set of estimates and leave more sophisticated designs for future work. To obtain each estimate, we employ a network with sufficient capacity to effectively model natural images. The architectural details are provided in Appendix~\ref{app:add_experiment}. Training the neural networks as in \eqref{eq:PGD} yields an estimate of the image from the noisy observation.
We reshape the vector $\xv^t - \mu_t \nabla f_L(\xv^t)$ into an image of size $(H \times W)$ and partition it into non-overlapping patches of size $(h_k \times w_k)$. Note that $h_k = H / k$, $w_k = W / k$, and $k \in \mathbb{N}$ is chosen such that $h_k \mid H$ and $w_k \mid W$, where $h_1 = H$ and $w_1 = W$. We then initialize independent DIPs, each with the same architecture as described above, to reconstruct the $(h_k \times w_k)$ patches individually. This results in $\frac{HW}{h_k w_k}$ separate DIP models. By placing the reconstructed patches back into their original spatial locations, we obtain an estimate of the entire image, denoted by $\check{\xv}_k$. A key property of this estimate is that each pixel depends only on the $(h_k \times w_k)$ patch to which it belongs and is independent of all other pixels. Repeating this procedure for $K$ different patch sizes yields $K$ estimates, denoted by $\check{\xv}_1, \ldots, \check{\xv}_K$. The final estimate is obtained by averaging these individual estimates.

Since the estimate $\check{\xv}_k$ for each $k$ relies on distinct local regions of the image, the corresponding pixel-wise estimates are expected to be weakly dependent across different values of $k$, in the sense that their cross-correlations are not close to $1$ or $-1$.

\subsubsection{Challenge 2: Matrix inversion}

The gradient of $f_L(\xv)$ with respect to the $j$-th coordinate of $\xv$, denoted by $x_j$, as used in PGD~\eqref{eq:PGD}, can be written as follows (see Appendix~\ref{app:MLE} for details)
\begin{align} \label{eq:GD-ML}
    \frac{\partial f_L}{\partial x_j}
    &=  2 \textcolor{blue}{x_j} \sigma_w^2 \left( \tilde{\av}_{\cdot, j}^{+T} B^{-1} \tilde{\av}_{\cdot, j}^{+} + \tilde{\av}_{\cdot, j}^{-T} B^{-1} \tilde{\av}_{\cdot, j}^{-} \right) 
    \notag\\
    \quad &- \frac{2 \xv_j \sigma_w^2}{L } \sum_{\ell=1}^L \left[ \left( \tilde{\av}_{\cdot, j}^{+T} B^{-1} \tilde{\yv}_\ell \right)^2 + \left( \tilde{\av}_{\cdot, j}^{-T} B^{-1} \tilde{\yv}_\ell \right)^2 \right],
\end{align}
where
$\Tilde{\mathbf{a}}^+_{\cdot, j} = \begin{bmatrix} \Re (\mathbf{a}_{\cdot, j}) \\ \Im (\mathbf{a}_{\cdot, j}) \end{bmatrix}$, $\Tilde{\mathbf{a}}^-_{\cdot, j} = \begin{bmatrix} -\Im (\mathbf{a}_{\cdot, j}) \\ \Re (\mathbf{a}_{\cdot, j}) \end{bmatrix}$, $\tilde{\yv}_\ell = \begin{bmatrix} \Re (\yv_\ell) \\ \Im (\yv_\ell) \end{bmatrix}$, $\mathbf{a}_{\cdot, j}$ denotes the $j$-th column of matrix $A$. It's important to highlight that, in each iteration $t$ of the PGD, the matrix $B$ as defined in~\eqref{eq:complex-cov} changes because it depends on the current estimate $\xv^t$. 
This leads to the computation of the inverse of a large matrix $B(\xv_t) \in \mathbb{R}^{2m \times 2m}$ at each iteration, posing a considerable computational challenge and a significant obstacle in applying DIP-based PGD for this problem. In the next section, we present a solution to address this issue.

\subsubsection{Solution to Challenge 2}

To address the challenge described in the previous section, we propose using the Newton-Schulz algorithm, an iterative method for approximating matrix inverses. Specifically, to compute $(B_t)^{-1}$, where we denote $B_t \triangleq B(\xv_t)$ for simplicity, the Newton-Schulz iterations are given by
\begin{equation}
	M^{\tilde k} = M^{\tilde k-1} + M^{\tilde k-1} \big(I - B_t M^{\tilde k-1}\big),
\end{equation}
where $M^{\tilde k}$ denotes the approximation of $(B_t)^{-1}$ at the $\tilde k$-th iteration. The initialization is set as $M^0 = (B_{t-1})^{-1}$, where $(B_{t-1})^{-1}$ is saved from the $(t-1)$-th PGD iteration. It is known that if $\sigma_{\max}(I - M^0 B_t) < 1$, then the Newton-Schulz iteration converges quadratically to $B_t^{-1}$~\cite{gower2017randomized, stotsky2020convergence}.

An observation to alleviate the issue mentioned in the previous section is that, given the nature of gradient descent, we don't anticipate significant changes in the matrix $X_t^2$ from one iteration to the next. Consequently, we expect $B_t$ and $B_{t-1}$, as well as their inverses, to be close to each other.
Hence, instead of calculating the full inverse at iteration $t$, we can employ the Newton-Schulz algorithm with $M^0$ set to $(B_{t-1})^{-1}$ from the previous iteration. 
}

\section{Simulation Results}

\label{sec:simulations}

\subsection{Study of the Impacts of Different Modules}

\subsubsection{Newton-Schulz iterations}

In this section, we aim to answer the following questions: (1) Is the Newton-Schulz algorithm effective in our Bagged-DIP-based PGD? (2) What is the minimum number of iterations for the Newton-Schulz algorithm to have good performance in Bagged-DIP-based PGD? (3) How does the computation time differ when using the Newton-Schulz algorithm compared to exact inverse computation? 
\begin{figure}[ht]
\centering
\includegraphics[width=0.47\textwidth]{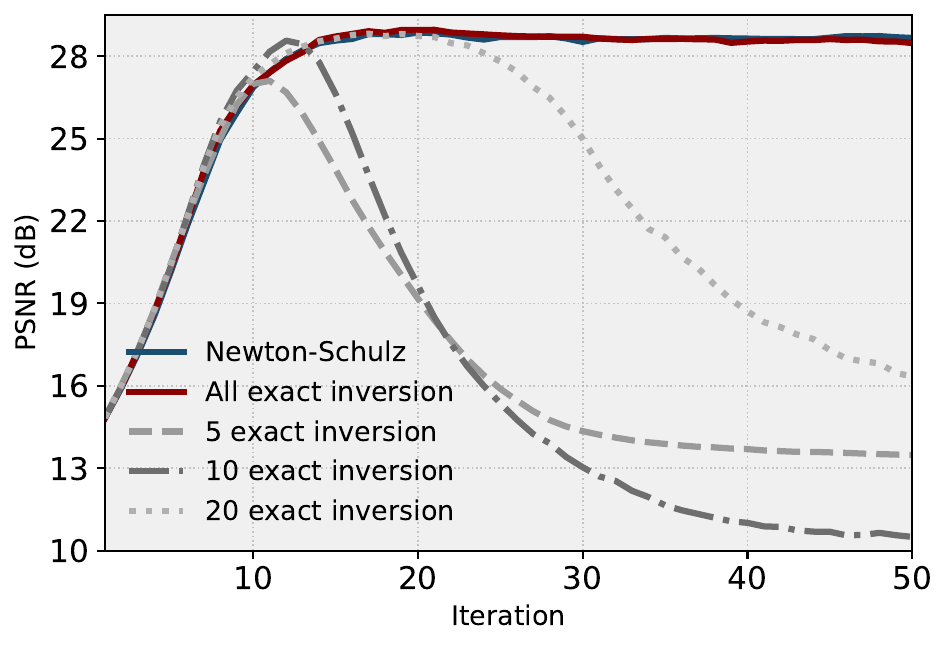}
\caption{Newton-Schulz approximation compared with computing the exact inverse for all iterations, the rest of the curves correspond to stopping the update of the inverse after the first $5$, $10$, and $20$ iterations respectively. The number of looks is $L=32$, sampling rate is $m/n=0.5$. The test image is ``Cameraman".}
\label{fig:iterative_alg}
\end{figure}

Figure \ref{fig:iterative_alg} shows one of the simulations we ran to address the first two questions. In this figure, we have chosen $L=32$ and $m/n =0.5$, and the learning rate of PGD is $0.01$. The result of Bagged-DIP-based PGD with a \textbf{single step} of Newton-Schulz is virtually identical to PGD with the exact inverse. To investigate the impact of the Newton-Schulz algorithm further, we next checked if applying even one step of Newton-Schulz is necessary. Hence, in three different simulations we stopped the matrix inverse update at iterations $5$, $10$, and $20$. As is clear from Figure \ref{fig:iterative_alg}, a few iterations after stopping the update, PGD starts diverging. Hence, we conclude that a single step of the Newton-Schulz is necessary and sufficient for PGD. 

To address the last question raised above, we evaluated how much time the calculation of the gradient takes if we use one step of the Newton-Schulz compared to the full matrix inversion. Our results are reported in Table \ref{tab:Newton_Schulz_time}. Our simulations are for sampling rate $50\%,$ and number of looks $ L=50$ and three different image sizes.\footnote{Our algorithm still faces memory limitations on a single GPU when processing $256 \times 256$ images. Addressing this issue through approaches like parallelization remains a subject for future research.} As is clear, the Newton-Schulz is much faster.  

\begin{table}[ht]
    \centering
    \caption{Runtime (in seconds) for matrix inversion and its Newton–Schulz approximation in each PGD step. Image sizes $32 \times 32$, $64 \times 64$, $128 \times 128$ are considered.}
    \label{tab:Newton_Schulz_time}
    \renewcommand{\arraystretch}{1.2}
    \begin{tabular}{lccc}
        \toprule
        \textbf{Method} & \textbf{32$\times$32} & \textbf{64$\times$64} & \textbf{128$\times$128} \\
        \midrule
        GD w/ Newton–Schulz & $\sim$7\text{e}{-5} & $\sim$8\text{e}{-5} & $\sim$1\text{e}{-4} \\
        GD w/o Newton–Schulz & $\sim$0.3 & $\sim$1.2 & $\sim$52.8 \\
        \bottomrule
    \end{tabular}
\end{table}

In our final algorithm, i.e. Bagged DIP-based PGD, if the difference between $\|\xv^{t}-\xv^{t-1}\|_{\infty} > \delta_{\xv}$, then we use the exact inverse update. $\delta_{\xv}$ is set to 0.12 (please refer to Appendix \ref{app:add_experiment} for details) in all our simulations. Based on this updating criterion, we observe that the exact matrix inverse is only required for the first 2-3 iterations, and it is adaptive enough to guarantee the convergence of PGD.

\subsubsection{Bagged-DIP}
\begin{figure}[t]
\centering
\includegraphics[width=0.47\textwidth]{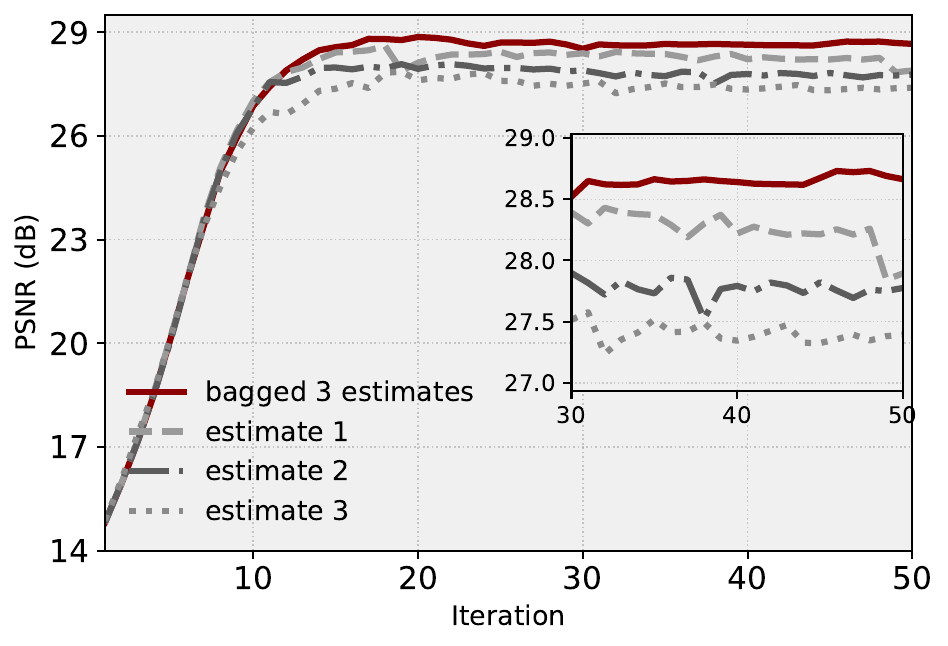}
\includegraphics[width=0.47\textwidth]{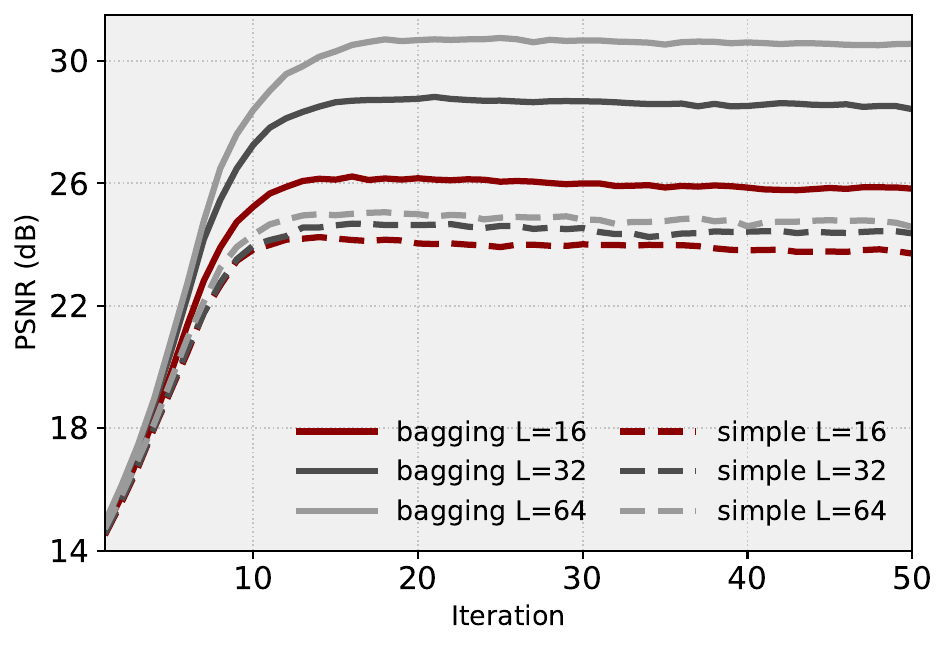}

\caption{(Left) We compare a Bagged-DIP with three sophisticated DIP estimates, where $L=32$, $m/n=0.5$. (Right) We compare PGD with simple and Bagged-DIP across different looks $L=16,32,64$. The test image is ``Cameraman".}
\label{fig:simple_bagged_L}
\end{figure}

Intuitively speaking, in bagging, the more weakly dependent estimates one generates, the better the average estimate will be. In the context of DIP, there appear to be many different ways to create weakly dependent samples. The goal of this section is not to explore the full-potential of Bagged-DIP. Instead, we aim to demonstrate that even a few weakly dependent samples can offer noticeable improvements. Hence, unlike the classical applications of bagging in which thousands of bagged samples are generated, to keep the computations manageable, we have only considered three bagged estimates. Figure \ref{fig:simple_bagged_L} shows one of our simulations. In this simulation, we have chosen $K=3$, i.e. we have only three weakly-dependent estimates. These estimates are constructed according to the recipe presented in Section \ref{ssec:baggedDIP} with the following patch sizes: 
$h_1 = w_1 = 32$, $h_2 = w_2 = 64$, and $h_3 = w_3 = 128$.
As is clear from the left panel of Figure \ref{fig:simple_bagged_L}, even with these very few samples, Bagged-DIP has offered between $0.5$dB and $1$dB  over the three estimates it has combined.

\subsubsection{Simple architectures versus Bagged-DIP}
So far our simulations have been focused on sophisticated networks. Are simpler networks that trade variance for the bias able to offer better performance? The right panel of Figure \ref{fig:simple_bagged_L} compares the performance of Bagged-DIP-based PGD with that of PGD with a simple DIP. Not only this figure shows the major improvement that is offered by using more complicated networks (in addition to bagging), but also it clarifies one of the serious limitations of the simple networks. Note that as $L$ increases, the performance of PGD with simple DIP is not improving. In such cases, the low capacity of DIP blocks the algorithm from taking advantage of extra information offered by the new looks.

\begin{table*}[t]
    \centering
    \caption{PSNR (dB) / SSIM $\uparrow$ of 8 test images recovered from measurements $\yv_\ell = A X \wv_\ell, \ell=1, \cdots, L$, at sampling rates $m/n = 0.125, 0.25, 0.5$, with number of looks $L = 1, 2, 4, 8, 16, 32, 64, 128$.}
    \label{tab:main-CS-PSNR-SSIM}
    \vspace{0.5em}

\resizebox{\textwidth}{!}{
\begin{tabular}{llccccccccc}
\toprule
\textbf{m/n} & \textbf{\#Looks} & \textbf{Barbara} & \textbf{Peppers} & \textbf{House} & \textbf{Foreman} & \textbf{Boats} & \textbf{Parrots} & \textbf{Cameraman} & \textbf{Monarch} & \textbf{Average} \\
\midrule
\multirow{8}{*}{0.125}
  & 1   & 13.05/0.126 & 13.34/0.132 & 12.90/0.127 & 11.20/0.083 & 12.92/0.098 & 10.40/0.090 &  9.08/0.069 & 12.36/0.161 & 11.91/0.111 \\
  & 2   & 14.62/0.179 & 15.03/0.175 & 15.33/0.190 & 12.83/0.131 & 15.54/0.161 & 12.35/0.146 & 10.49/0.112 & 14.15/0.241 & 13.79/0.167 \\
  & 4   & 16.11/0.273 & 16.80/0.322 & 16.49/0.274 & 14.29/0.268 & 17.32/0.247 & 13.64/0.228 & 11.75/0.156 & 15.60/0.327 & 15.25/0.262 \\
  & 8   & 16.84/0.376 & 17.37/0.465 & 17.51/0.429 & 14.63/0.446 & 18.50/0.356 & 14.19/0.311 & 12.26/0.203 & 16.18/0.419 & 15.94/0.376 \\
  & 16  & 17.78/0.293 & 18.11/0.305 & 18.20/0.286 & 17.98/0.257 & 17.86/0.259 & 15.36/0.279 & 13.01/0.226 & 17.26/0.463 & 16.95/0.296 \\
  & 32  & 20.88/0.462 & 20.41/0.438 & 21.55/0.445 & 19.64/0.448 & 20.74/0.398 & 18.11/0.411 & 15.68/0.318 & 19.78/0.570 & 19.60/0.436 \\
  & 64  & 21.71/0.587 & 21.92/0.585 & 23.52/0.571 & 20.80/0.590 & 22.21/0.503 & 19.27/0.522 & 17.73/0.407 & 21.72/0.675 & 21.11/0.555 \\
  & 128 & 22.67/0.647 & 22.70/0.667 & 24.09/0.651 & 21.07/0.684 & 22.33/0.550 & 19.42/0.587 & 19.06/0.506 & 22.77/0.740 & 21.76/0.629 \\
\midrule
\multirow{8}{*}{0.25}
  & 1   & 14.04/0.157 & 14.41/0.159 & 14.01/0.146 & 12.61/0.100 & 14.24/0.123 & 11.50/0.125 &  9.97/0.117 & 13.28/0.280 & 13.01/0.151 \\
  & 2   & 16.23/0.235 & 16.85/0.245 & 17.17/0.277 & 14.74/0.195 & 16.89/0.208 & 13.87/0.226 & 11.96/0.189 & 16.02/0.392 & 15.47/0.246 \\
  & 4   & 18.25/0.388 & 18.78/0.394 & 18.67/0.368 & 16.25/0.321 & 18.73/0.313 & 15.31/0.323 & 13.82/0.256 & 17.84/0.504 & 17.21/0.358 \\
  & 8   & 18.94/0.480 & 19.41/0.516 & 20.02/0.530 & 17.50/0.544 & 19.90/0.430 & 16.57/0.436 & 15.09/0.328 & 18.84/0.591 & 18.28/0.482 \\
  & 16  & 21.76/0.491 & 21.41/0.445 & 21.70/0.417 & 21.01/0.385 & 20.79/0.398 & 19.25/0.429 & 19.13/0.417 & 20.55/0.612 & 20.70/0.449 \\
  & 32  & 23.96/0.611 & 24.05/0.600 & 25.17/0.573 & 23.88/0.593 & 23.13/0.528 & 22.18/0.571 & 22.33/0.526 & 23.29/0.713 & 23.50/0.590 \\
  & 64  & 25.81/0.734 & 25.74/0.719 & 27.66/0.683 & 25.27/0.714 & 24.84/0.632 & 23.99/0.671 & 25.04/0.656 & 25.23/0.804 & 25.45/0.702 \\
  & 128 & 26.70/0.774 & 26.34/0.782 & 28.76/0.741 & 26.54/0.791 & 25.59/0.683 & 24.80/0.741 & 27.01/0.780 & 26.44/0.857 & 26.52/0.769 \\
\midrule
\multirow{8}{*}{0.5}
  & 1   & 16.31/0.242 & 16.77/0.246 & 16.42/0.225 & 15.70/0.164 & 16.36/0.213 & 13.87/0.255 & 14.00/0.317 & 15.73/0.413 & 15.65/0.259 \\
  & 2   & 19.17/0.388 & 19.70/0.376 & 19.57/0.360 & 17.61/0.297 & 19.26/0.315 & 17.03/0.381 & 18.23/0.503 & 19.07/0.556 & 18.70/0.397 \\
  & 4   & 21.03/0.533 & 21.57/0.549 & 22.03/0.518 & 20.08/0.491 & 21.74/0.456 & 19.32/0.515 & 21.28/0.639 & 21.78/0.683 & 21.10/0.548 \\
  & 8   & 22.19/0.650 & 22.73/0.666 & 23.84/0.639 & 21.51/0.683 & 22.72/0.565 & 20.22/0.620 & 24.27/0.773 & 23.08/0.760 & 22.57/0.669 \\
  & 16  & 25.85/0.691 & 25.67/0.636 & 26.53/0.609 & 25.77/0.621 & 24.83/0.596 & 24.52/0.664 & 25.87/0.675 & 25.48/0.789 & 25.56/0.660 \\
  & 32  & 27.77/0.780 & 27.70/0.753 & 29.26/0.725 & 27.76/0.758 & 26.81/0.699 & 26.23/0.752 & 28.47/0.778 & 27.84/0.868 & 27.73/0.764 \\
  & 64  & 28.91/0.823 & 28.70/0.825 & 30.90/0.790 & 28.99/0.842 & 27.98/0.763 & 27.64/0.822 & 30.47/0.859 & 29.30/0.914 & 29.11/0.830 \\
  & 128 & 29.43/0.846 & 29.37/0.861 & 31.71/0.824 & 29.33/0.885 & 28.57/0.792 & 28.28/0.859 & 31.76/0.911 & 30.07/0.937 & 29.82/0.864 \\
\bottomrule
\end{tabular}
}
\end{table*}

\subsection{Performance of Bagged-DIP-based PGD}
\label{sec:simulation_baggedDIP}

In this section, we offer a comprehensive simulation study to evaluate the performance of the Bagged-DIP-based PGD on several images. We explore the following settings in our simulations:

\begin{itemize}
\item Number of looks ($L$): $L=1 ,2, 4, 8, 16, 32, 64, 128$.\

\item Undersampling rate ($m\over n$): $\frac{m}{n}= 0.125, 0.25, 0.5$. \

\end{itemize}

For each combination of $L$ and $m/n$, we pick one of the $8$, $128 \times 128$ images mentioned in Table \ref{tab:main-CS-PSNR-SSIM}.\footnote{Images from the Set11~\cite{kulkarni2016reconnet} are chosen and cropped to $128 \times 128$ for computational manageability in Table~\ref{tab:main-CS-PSNR-SSIM}.} We then generate the matrix $A \in \mathbb{C}^{m \times n}$ by selecting the first $m$ rows of a matrix that is drawn from the Haar measure on the space of orthogonal matrices.  We then generate $\wv_1, \ldots, \wv_L \sim \mathcal{CN} (0,1)$, and for $\ell=1, 2, \ldots, L$, calculate $\yv_\ell = A X_o \wv_\ell$.

For our implementation of Bagged-DIP-based PGD, we have made the following choices:

\begin{itemize}
\item Initialization: We initialize our algorithm with ${\bx}_0 = \frac{1}{L} \sum^L_{\ell=1} |\Bar{A}^{\top} \yv_\ell|$. However, the final performance of DIP-based PGD is robust to the choice of  initialization. 

\item Learning rate: We have selected a learning rate of $0.001$ for the gradient descent of the likelihood when $L \leq 8$, and $0.01$ otherwise, and learning rate of $0.001$ in the training of DIP.

\item Number of iterations of SGD for training DIP: The details are presented in Table~\ref{tab:num_iteration_DIP} in the appendix.

\item Number of iterations of PGD: We run the outer loop (gradient descent of likelihood) for $100,200,300$ iterations when $m/n=0.5,0.25,0.125$ respectively.
\end{itemize}

The peak signal-to-noise-ratio (PSNR) and structural index similarity (SSIM) of our reconstructions are all reported in Table~\ref{tab:main-CS-PSNR-SSIM}. {The reconstructed images are exhibited in Figures \ref{fig:vis_S125},\ref{fig:vis_S25},\ref{fig:vis_S50} in Appendix~\ref{app:visualization}}

{\subsection{Effect of additive noise on the recovery performance}
In this section, we study the effect of additive white Gaussian noise on the performance of our algorithm. Towards this goal, we consider the measurements $\yv_\ell = A X_o \wv_\ell + \zv_\ell$, where $\zv_1, \ldots, \zv_L \sim \mathcal{CN} (0,\sigma^2_z I_m)$, $\wv_1, \ldots, \wv_L \sim \mathcal{CN} (0,I_n)$, for $\ell=1, 2, \ldots, L$. We consider four different values for $\sigma_z$. These values are shown in Table \ref{tab:main-CS-PSNR-SSIM-awgn}. Since the measurements are complex-valued and additive noise exists in the measurements, we use the formulations presented in \eqref{eq:complex-likelihood}. The recovery results for different levels of additive noise are reported in Table \ref{tab:main-CS-PSNR-SSIM-awgn}. As is clear from the results reported in this table, the effect of additive noise level $\sigma_z$ on the performance of our algorithm is minor. This is consistent with the conventional understanding in coherent imaging systems that speckle noise has the dominant impact on reconstruction accuracy, while the effect of additive noise can often be neglected in practice. 
}

\begin{table*}[t]
    \centering
    \caption{PSNR (dB) / SSIM $\uparrow$ of 8 test images when additive noise levels in the measurement model $\yv_\ell = AX \wv_\ell + \zv_\ell$ are $\sigma_z = 0, 15, 25, 50$, with number of looks $L = 16, 32, 64, 128$, sampling rate $m/n=0.25$. The recovery algorithm considers the additive noise.}
    \label{tab:main-CS-PSNR-SSIM-awgn}
    \vspace{0.5em}
\resizebox{\textwidth}{!}{
\begin{tabular}{llccccccccc}
\toprule
 \textbf{\#Looks} & \textbf{$\sigma_z$} & \textbf{Barbara} & \textbf{Peppers} & \textbf{House} & \textbf{Foreman} & \textbf{Boats} & \textbf{Parrots} & \textbf{Cameraman} & \textbf{Monarch} & \textbf{Average} \\
\midrule
\multirow{4}{*}{16}
  & 0  & 21.76/0.491 & 21.41/0.445 & 21.70/0.417 & 21.01/0.385 & 20.79/0.398 & 19.25/0.429 & 19.13/0.417 & 20.55/0.612 & 20.70/0.449 \\
  & 15  & 21.17/0.456 & 21.17/0.422 & 21.70/0.416 & 21.00/0.370 & 20.62/0.401 & 19.19/0.439 & 18.86/0.407 & 20.37/0.606 & 20.51/0.440 \\
  & 25  & 20.99/0.447 & 21.14/0.418 & 21.77/0.419 & 20.91/0.368 & 20.46/0.397 & 18.99/0.431 & 18.39/0.395 & 20.29/0.604 & 20.37/0.435 \\
  & 50 & 20.80/0.432 & 20.66/0.398 & 21.25/0.398 & 20.74/0.356 & 20.33/0.380 & 18.22/0.403 & 16.91/0.353 & 19.60/0.577 & 19.81/0.412 \\
\midrule
\multirow{4}{*}{32}
  & 0  & 23.96/0.611 & 24.05/0.600 & 25.17/0.573 & 23.88/0.593 & 23.13/0.528 & 22.18/0.571 & 22.33/0.526 & 23.29/0.713 & 23.50/0.590 \\
  & 15  & 24.12/0.631 & 23.82/0.586 & 25.07/0.563 & 23.86/0.587 & 23.54/0.522 & 22.01/0.564 & 22.38/0.533 & 23.29/0.715 & 23.51/0.588 \\
  & 25  & 24.00/0.625 & 23.81/0.582 & 24.85/0.556 & 23.93/0.586 & 23.50/0.523 & 21.96/0.562 & 21.91/0.509 & 23.06/0.709 & 23.38/0.581 \\
  & 50  & 23.19/0.596 & 23.32/0.560 & 24.33/0.545 & 23.79/0.584 & 23.21/0.506 & 21.21/0.538 & 19.78/0.439 & 22.34/0.680 & 22.64/0.556 \\
\midrule
\multirow{4}{*}{64}
  & 0  & 25.81/0.734 & 25.74/0.719 & 27.66/0.683 & 25.27/0.714 & 24.84/0.632 & 23.99/0.671 & 25.04/0.656 & 25.23/0.804 & 25.45/0.702 \\
  & 15  & 26.35/0.721 & 25.56/0.713 & 27.56/0.676 & 25.76/0.727 & 25.28/0.636 & 23.76/0.665 & 24.61/0.629 & 25.25/0.805 & 25.52/0.697 \\
  & 25  & 26.13/0.715 & 25.40/0.709 & 27.47/0.671 & 25.11/0.720 & 25.32/0.632 & 23.50/0.658 & 24.30/0.619 & 25.13/0.800 & 25.30/0.690 \\
  & 50  & 25.48/0.695 & 24.79/0.692 & 26.87/0.654 & 24.07/0.693 & 24.50/0.613 & 22.94/0.639 & 21.80/0.541 & 24.21/0.773 & 24.33/0.662 \\
\midrule
\multirow{4}{*}{128}
  & 0  & 26.70/0.774 & 26.34/0.782 & 28.76/0.741 & 26.54/0.791 & 25.59/0.683 & 24.80/0.741 & 27.01/0.780 & 26.44/0.857 & 26.52/0.769 \\
  & 15  & 27.04/0.767 & 26.37/0.779 & 28.77/0.744 & 25.61/0.780 & 25.32/0.678 & 24.32/0.735 & 26.24/0.745 & 26.25/0.855 & 26.24/0.760 \\
  & 25  & 26.68/0.760 & 26.39/0.776 & 28.68/0.741 & 25.33/0.777 & 25.33/0.675 & 24.35/0.732 & 26.20/0.752 & 26.19/0.849 & 26.14/0.758 \\
  & 50  & 26.08/0.744 & 25.64/0.761 & 28.24/0.731 & 25.86/0.775 & 25.51/0.665 & 23.36/0.706 & 23.34/0.660 & 25.36/0.823 & 25.42/0.733 \\
\bottomrule
\end{tabular}
}
\end{table*}

{
\subsection{An empirical upper bound}
As we described before, the existing literature cannot be directly applied to the undersampled forward models, i.e., the case $m<n$, considered in this paper. As a result, we are unable to make a direct comparison between our algorithm and existing approaches. Nevertheless, it remains valuable to compare our method with neural network–based recovery algorithms.
Hence, towards this goal, we consider the regime $m=n$, i.e., when the number of measurements equals the signal ambient dimension. Under this setting, we first show the results of our algorithm in Table~\ref{tab:oracle-PSNR-SSIM}, with measurement model $\yv_\ell = A X_o \wv_\ell + \zv_\ell$, where $m/n=1.0$, and $\sigma_z=25$. Note that in the table we have considered $8$ different number of looks. 

\begin{figure}[h]
\centering
\includegraphics[width=0.47\textwidth]{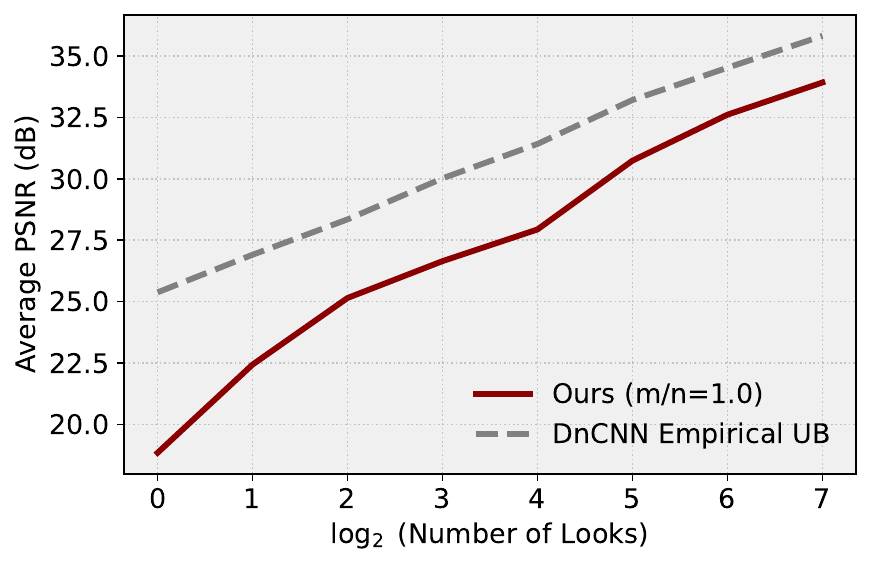}
\caption{We compare the average PSNR reconstruction performance on 8 test images of our algorithm and the supervised trained DnCNN, when the measurement model is $\yv_\ell = AX \wv_\ell + \zv_\ell$, where sampling rate $m/n=1.0$, additive noise level $\sigma_z=25$.}
\label{fig:UB}
\end{figure}

To obtain an empirical upper bound (UB), we adopt the same measurement model as before, with sampling rate $m/n = 1.0$ and noise level $\sigma_z = 25$. However, we assume that we have access to a training set that includes the “underlying images” and their corresponding measurements through a coherent imaging system. Using this extra information, we train a DnCNN model \cite{zhang2017beyond} that receives $\sqrt{\frac{1}{L} \sum_{\ell=1}^{L} |\bar{A}^{\top} \yv_\ell|^2}$ as its input and aims to reconstruct $\xv_o$. We have chosen the images from the BSD400 dataset \cite{martin2001database}. The trained network is then used to reconstruct test images from their measurements. We refer to this approach as an empirical “upper bound (UB)” since the reconstruction relies on a high-capacity neural network trained in an end-to-end supervised manner on a sufficiently large dataset. Note that our Bagged-DIP PGD algorithm does not require any training set. The corresponding results are reported in Table~\ref{tab:oracle-PSNR-SSIM}, with additional implementation details provided in Appendix~\ref{app:empirical_UB}.

A general pattern that is observed is that when the number of looks is small, e.g., $L=1$ or $L=2$, the difference between the performance of the empirical UB and our algorithm is significant ($\sim 8$dB in PSNR). However, as the number of looks increases, the gain starts to shrink, for instance, when $L=128$, the gain in the PSNR reduces to $1$dB. In Figure~\ref{fig:UB}, we further compare the empirical UB achieved by DnCNN with the average PSNR of the images reconstructed by our algorithm across different numbers of looks.}

\begin{table*}[t]
    \centering
    \caption{PSNR (dB) / SSIM $\uparrow$ of 8 test images recovered from measurements $\yv_\ell = A X \wv_\ell + \zv_\ell, \ell=1, \cdots, L$, at sampling rate $m/n = 1.0$, with additive noise level $\sigma_z = 25$. Furthermore, we provide an empirical upper bound, where measurement model is the same as before, the images are recovered by a DnCNN trained on pairs of images $(\xv, \sqrt{\frac{1}{L} \sum^L_{\ell=1} |\bar{A}^{\top} \yv_\ell|^2})$.}
    \label{tab:oracle-PSNR-SSIM}
    \vspace{0.5em}

\resizebox{\textwidth}{!}{
\begin{tabular}{llccccccccc}
\toprule
\textbf{m/n} & \textbf{\#Looks} & \textbf{Barbara} & \textbf{Peppers} & \textbf{House} & \textbf{Foreman} & \textbf{Boats} & \textbf{Parrots} & \textbf{Cameraman} & \textbf{Monarch} & \textbf{Average} \\
\midrule
\multirow{8}{*}{1.0}
  & 1   & 20.01/0.425 & 20.01/0.387 & 19.85/0.369 & 20.16/0.334 & 19.25/0.368 & 17.30/0.442 & 15.46/0.482 & 18.73/0.579 & 18.85/0.423 \\
  & 2   & 23.72/0.616 & 23.47/0.536 & 23.57/0.509 & 24.12/0.547 & 22.23/0.480 & 21.20/0.591 & 18.79/0.589 & 22.40/0.714 & 22.44/0.572 \\
  & 4   & 25.99/0.718 & 25.75/0.675 & 26.89/0.654 & 27.11/0.737 & 24.73/0.595 & 23.80/0.706 & 22.21/0.684 & 24.72/0.803 & 25.15/0.696 \\
  & 8   & 27.11/0.772 & 27.16/0.769 & 28.53/0.744 & 28.13/0.821 & 26.37/0.691 & 24.95/0.781 & 24.86/0.765 & 26.03/0.863 & 26.65/0.776 \\
  & 16  & 28.32/0.779 & 28.13/0.705 & 28.53/0.685 & 29.93/0.713 & 27.56/0.700 & 27.10/0.784 & 25.95/0.733 & 27.95/0.849 & 27.94/0.743 \\
  & 32  & 30.98/0.862 & 30.51/0.816 & 31.80/0.803 & 33.12/0.803 & 30.58/0.813 & 29.58/0.864 & 28.84/0.813 & 30.51/0.910 & 30.74/0.840 \\
  & 64  & 32.45/0.906 & 32.12/0.883 & 33.76/0.862 & 35.01/0.901 & 32.65/0.874 & 31.16/0.911 & 31.04/0.874 & 32.71/0.947 & 32.61/0.895 \\
  & 128 & 33.32/0.925 & 32.97/0.915 & 35.07/0.891 & 36.64/0.938 & 34.11/0.913 & 32.24/0.932 & 32.93/0.939 & 34.04/0.967 & 33.92/0.928 \\
\midrule
\multirow{8}{*}{\shortstack{Empirical\\ ``UB"}}
  & 1 & 25.07/0.661 & 26.30/0.769 & 26.24/0.725 & 28.80/0.815 & 25.05/0.641 & 23.58/0.733 & 24.08/0.852 & 23.92/0.818 & 25.38/0.752 \\
  & 2 & 26.62/0.730 & 27.39/0.796 & 27.88/0.768 & 29.72/0.844 & 26.53/0.705 & 25.13/0.771 & 26.01/0.885 & 26.03/0.876 & 26.91/0.797 \\
  & 4 & 27.96/0.792 & 28.93/0.830 & 29.21/0.788 & 30.47/0.839 & 27.90/0.761 & 26.73/0.807 & 27.97/0.908 & 27.59/0.900 & 28.35/0.828 \\
  & 8 & 29.47/0.844 & 30.38/0.861 & 31.12/0.819 & 32.36/0.879 & 29.57/0.809 & 28.20/0.842 & 29.51/0.922 & 29.56/0.933 & 30.02/0.864 \\
  & 16 & 30.83/0.875 & 31.86/0.890 & 32.15/0.833 & 33.03/0.898 & 31.01/0.845 & 29.43/0.873 & 31.63/0.945 & 31.39/0.955 & 31.42/0.889 \\
  & 32 & 32.72/0.913 & 33.40/0.911 & 33.45/0.859 & 35.59/0.916 & 32.53/0.887 & 31.03/0.904 & 33.67/0.953 & 33.32/0.966 & 33.21/0.914 \\
  & 64 & 34.14/0.932 & 34.77/0.930 & 35.02/0.891 & 36.05/0.928 & 33.89/0.913 & 32.15/0.921 & 35.04/0.964 & 35.11/0.976 & 34.52/0.932 \\
  & 128 & 36.13/0.956 & 36.11/0.943 & 35.75/0.901 & 37.43/0.932 & 35.55/0.940 & 33.87/0.950 & 35.31/0.962 & 36.46/0.983 & 35.82/0.946 \\
\bottomrule
\end{tabular}
}
\end{table*}


\section{Proof of \prettyref{thm:maintheorem}}
\label{sec:proof-theory}

\subsection{Auxiliary lemmas}

We will use the following technical results from the literature to prove \prettyref{thm:maintheorem}.

\begin{lemma}\label{lemma:1} \cite{zhou2022compressed}
	Let $\xvt \in\mathbb{R}^n$ and  $\xv_o\in\mathbb{R}^n$ and assume that $A\tilde{X}^2A^{\top}$ and $AX_o^2A^{\top}$ are both invertible. Define $\tilde{X},X_o,\tilde{\Sigma}, \Sigma_o,\Delta\tilde \Sigma$ as in \eqref{eq:proof-notations}.
	Let $\lambda_{\max}$ be the operator norm of $\Sigma_o^{-\frac 12}\Delta\tilde \Sigma \Sigma_o^{-\frac 12}$.  Then,
	\begin{align*}
		\bar{f}(\tilde{\Sigma})-\bar{f}(\Sigma_o)&\geq {1\over 2(1+\lambda_{\max})^2}{\rm Tr}(\Sigma_o^{-1}\Delta\tilde \Sigma\Sigma_o^{-1}\Delta \tilde \Sigma).
	\end{align*}
\end{lemma}

\begin{lemma}\label{lem:singvalues}\cite{RudelsonVershinin2010}
	Let the elements of an $m \times n$ ($m<n$) matrix $A$ be drawn independently from $\Nc(0,1)$. Then, for any $t>0$,
	\begin{align*}
		&\Prob(\sqrt{n}-\sqrt{m}- t \leq \sigma_{\min} (A) \leq \sigma_{\max}(A) \leq \sqrt{n}+\sqrt{m}+ t) 
		\notag\\
		&\geq 1-  2 e^{-\frac{t^2}{2}}.
	\end{align*}
\end{lemma}

\begin{lemma}\label{lemma:vector} \cite{zhou2022compressed}
	Consider the matrices $\tilde{\Sigma},\Sigma,\Delta \tilde \Sigma$ defined in \eqref{eq:proof-notations}. Then,
	\begin{align*}
		&\Tr(\Sigma^{-1}\Delta \tilde \Sigma\Sigma^{-1}\Delta \tilde\Sigma) 
		\geq {x^4_{\min} \lambda^2_{\min}(AA^{\top}) \over x^8_{\max} \lambda^4_{\max}(AA^{\top})  }\|A(\tilde{X}^2-X^2)A^{\top}\|^2_{\sf HS},\\
		&\Tr(\Sigma^{-1}\Delta \tilde \Sigma\Sigma^{-1}\Delta \tilde\Sigma)
		\leq {x^4_{\max} \lambda^2_{\max}(AA^{\top}) \over x_{\min}^8 \lambda^4_{\min}(AA^{\top})  }\|A(\tilde{X}^2-X^2)A^{\top}\|^2_{\sf HS}.   
	\end{align*}
\end{lemma}

We also use the following new technical results (proven in \prettyref{app:technical}) in the proof of \prettyref{thm:maintheorem}.

\begin{lemma}\label{lemma:concent-z}
	Consider fixed vectors $\xv_o,\xvt\in \reals^n$ such that $AX_o^2A^\top,A\tilde X^2A^\top$ are invertible, and define $\tilde \Sigma,\Sigma_o,\Delta\tilde\Sigma$ as in \eqref{eq:proof-notations}. Then, for $t>0$, there exists a constant $c$ independent of $m,n,L,k$, such that
	\begin{align*}
		&~\PP\pth{\left.\Biggl|{1\over L \sigma_w^2} \sum_{\ell=1}^L \yv_{\ell}^{\top} \Delta\tilde \Sigma \yv_{\ell}-{\rm Tr}(\Delta\tilde \Sigma \Sigma_o^{-1})\Biggr|\geq t\right |A}\\
		&\leq  2\exp\Biggl( - c\cdot \min\Biggl\{{Lt^2\over \Tr({\Sigma}_o^{-1}{\Delta \tilde \Sigma}{\Sigma}_o^{-1}{\Delta \tilde \Sigma})},
		 {\frac{Lt \cdot  x^4_{\min}  (\smin(A))^4}{x_{\max}^2(\sigma_{\max}(A))^4 \|{\xv}_o^2 - {\xvt}^2\|_\infty}}\Biggr\} \Biggr).
	\end{align*}
\end{lemma}

\begin{lemma}\label{lem:delta-error}
	There exist constants $\tilde{C},\bar C>0$ (which may depend on $x_{\min},x_{\max}$) such that the following holds. Suppose $\hat \bx = g_{\hat\btheta}(\bu)$ and $\tilde \bx = g_{\tilde\btheta}(\bu)$ with $\|\hat \btheta-\tilde\btheta\|_2\leq \delta$. Define $h(\Sigma)=\frac 1{L\sigma_w^2} \sum_{\ell=1}^L \yv_\ell^{\top} {(\Sigma-\Sigma_o)}\yv_\ell-\Tr((\Sigma-\Sigma_o){\Sigma}_o^{-1})$ and denote $\Sigma_o,\hat\Sigma,\tilde \Sigma,\Delta\Sigma,\Delta\tilde\Sigma$ as in \eqref{eq:proof-notations}. Then we have
	\begin{enumerate}[label=(\alph*.)]
		\item 
		$\PP\qth{|h(\tilde{\Sigma}_o)-h(\hat{\Sigma}_o) | \leq \tilde{C} n \delta}
		\geq 1 - O(e^{-\xi^2\frac{m}{2}} + e^{-\frac{Ln}{8}})$.
		
		\item 
		\begin{align*}
		&\PP\Biggl[\abs{\sqrt{\Tr({\Sigma}_o^{-1}\Delta{\tilde \Sigma}{\Sigma}_o^{-1}\Delta{\tilde \Sigma})}-\sqrt{\Tr({\Sigma}_o^{-1}\Delta{ \Sigma}{\Sigma}_o^{-1}\Delta{\tilde \Sigma})}}
		\quad \leq \sqrt{\bar C n\delta}\Biggr]\geq 1-2e^{-\xi^2\frac m2}
		\end{align*}.
	\end{enumerate}
\end{lemma}

{ \begin{lemma} \label{lem:lowerAX2AT}
	Let the elements of $m\times n$ matrix $A$ be drawn  i.i.d. $\calN(0,1)$. Define $q_{m,n} = m^2( 2\sqrt{n} + \sqrt{m})^4, 
	\tilde{q}_{m,n}= (2\sqrt{n} +\sqrt{m})^2$ and $D = \diag ({\bd})$ for any given ${\bd} \in \mathbb{R}^n$. Then, there exist absolute constants $c,C$ such that the following holds:
	\begin{align*}
		&\Prob (\|A DA^{\top}\|^2_{\sf HS} \leq m(m-1) \sum_{i=1}^n d_i^2 - t )
		\notag\\
		&\leq 2mC \exp \left( -c \min \Big( \frac{t^2}{C^2   \| {\bd}\|_{\infty}^{{4}}  q_{m,n} }, \frac{t}{C \| {\bd}\|_{\infty}^{2} \tilde{q}_{m,n}} \Big)\right) + 2 me^{-\frac{n}{2}}.
	\end{align*}
\end{lemma}}

\subsection{Details of proof of \prettyref{thm:maintheorem}}
We break the proof into several steps as follows. 
\subsubsection{Lower bounding $\bar f(\hat{\Sigma})-\bar f({\Sigma}_o)$}
We will establish that there is a constant $\tilde c>0$ such that
\begin{align}\label{eq:bound1}
	\bar{f}(\hat{\Sigma})-\bar f({\Sigma}_o)
	\geq
	\frac {\Tr({\Sigma}_o^{- 1}\Delta{\Sigma}{\Sigma}_o^{-1}\Delta{\Sigma})}{2(1+\tilde c)^2}
\end{align}
with a high probability. To show the above, we will first establish the inequality
\begin{equation}\label{eq:average:likelihood:lower}
	\bar{f}(\hat{\Sigma})-\bar f({\Sigma}_o)
	\geq \frac {\Tr({\Sigma}_o^{- 1}\Delta{\Sigma}{\Sigma}_o^{-1}\Delta{\Sigma})}{2(1+\lambda_{\max})^2},
\end{equation}
where $\lambda_{\max}$ denotes the operator norm of ${\Sigma}_o^{-\frac 12}\Delta{\Sigma}{\Sigma}_o^{-\frac 12}$. The above inequality follows directly from Lemma \ref{lemma:1}.
Consider substituting \(\tilde \Sigma\) and \(\Delta\tilde \Sigma\) with \(\hat \Sigma\), and   \(\Delta\Sigma\), respectively, in Lemma \ref{lemma:1}.  The lemma applies as long as both \(A\hat X^2A^\top\) and \(AX_o^2A^\top\) are invertible. Since \(AX^2A^\top\geq x_{\min}^2\sigma_{\min}(A)^2\) for \(X=X_o,\hat X\), the invertibility condition is satisfied on the event \(\calE_\xi\), defined below:
\begin{align}\label{eq:E4}
	\calE_\xi
	&=\{\sqrt n-(1+\xi)\sqrt m\leq \sigma_{\min}(A)\leq \sigma_{\max}(A)
	\notag\\
	&\leq \sqrt n+(1+\xi)\sqrt m\},
\end{align} 
for a small constant $\xi>0$ that satisfies $(1+\xi)^2 < 1+\xi_0$, with $\xi_0$ as in \prettyref{asmp:sensor-signal}. This ensures 
$$
\sqrt n-(1+\xi)\sqrt m
\geq 
\sqrt n-{1+\xi\over \sqrt {1+\xi_0}}\sqrt n
\geq c_0\sqrt{n}
$$ 
for a small constant $c_0>0$. \prettyref{lem:singvalues} establishes that $\calE_\xi$ is a high probability event.
That is, $\PP[\calE_\xi]\geq 1-2e^{-\xi^2\frac m2}$. 

Next, to establish \eqref{eq:bound1} from \eqref{eq:average:likelihood:lower}, we need to find a data-independent upper bound for $\lambda_{\max}$. We have
\begin{align}\label{eq:m1}
	&\lambda_{\max}
	=\max_{\uv\in\mathbb{R}^n}{ \|\Sigma_o^{-\frac 12} \Delta \Sigma \Sigma_o^{-\frac 12} \uv\|_2 \over \|\uv\|_2} 
	\notag\\
	&=\max_{\uv\in\mathbb{R}^n}{ \| \pth{\Sigma_o^{-\frac 12} \hat \Sigma \Sigma_o^{-\frac 12}-I} \uv\|_2 \over \|\uv\|_2}
	\leq \abs{\lambda_{\max}(\Sigma_o^{-1})\lambda_{\max}(\hat \Sigma)}+1.
\end{align}
Using the fact that $\xv_o,\xvh$ are coordinate-wise bounded from below by $x_{\min}$, we get from \eqref{eq:proof-notations}
$$
\lambda_{\max}(\hat \Sigma)
=
(\lambda_{\min}(A\hat X^2A^{\top}))^{-1}\leq (\lambda_{\min}(AA^{\top})x^2_{\min})^{-1},
$$ 
and
\begin{align*}
	\lambda_{\max}\pth{\Sigma_o^{-1}}
	=\lambda_{\max}(AX_o^2A^{\top})
	\leq \lambda_{\max}(AA^{\top}) x_{\max}^2.
\end{align*}
Hence, conditioned on $\calE_\xi$, using the inequality that for any matrix $A\in \reals^{m\times n}$ with $m\leq n$ we have
\begin{align}\label{eq:eigen-singular}
	\lambda_{\min}(AA^\top)\geq \sigma_{\min}(A)^2,
	\quad  \lambda_{\max}(AA^\top)\leq \sigma_{\max}(A)^2,
\end{align}
we continue \eqref{eq:m1} to get
\begin{align}\label{eq:upper:lambda:max}
	\lambda_{\max}
	&\leq {\lambda_{\max}(AA^{\top})x_{\max}^2\over
		\lambda_{\min}(AA^{\top})x_{\min}^2} + 1
	\notag\\
	&\leq {\pth{\sqrt n+(1+\xi)\sqrt m}^2x_{\max}^2\over \pth{\sqrt n-(1+\xi)\sqrt m}^2x_{\min}^2}+1
	\leq \tilde c,
\end{align}
for a constant $\tilde c>0$ depending on $\xi,x_{\min},x_{\max}$, whenever $m\leq {n\over (1+\xi)^2}$. This concludes the proof of \prettyref{eq:bound1}.

\subsubsection{Upper bounding $-\qth{\frac 1{L\sigma_w^2} \sum_{\ell=1}^L \texorpdfstring{\yv}{by}_\ell^{\top} \Delta{\Sigma} \texorpdfstring{\yv}{by}_\ell-\Tr(\Delta{\Sigma}{\Sigma}_o^{-1})}$}


We would like to substitute $\xvt$ with the observation-dependent random estimate $\xvh$ (i.e., $\Delta \tilde \Sigma$ is substituted with $\Delta \Sigma=\hat\Sigma-\Sigma_o$) in the above result to achieve the desired bound. However, the main challenge is that $\xvh$ is not a fixed vector as the above result requires. We use a covering set argument to circumnavigate this issue. In other words, we consider a $\delta$-net of the set $B_{\infty,k} (\mathbf{0}, x_{\max} \sqrt{n \over k})$ in terms of the Euclidean metric (i.e., given any $\thetav\in B_{\infty,k} (\mathbf{0}, x_{\max} \sqrt{n \over k})$, there is a $\tilde {\thetav}$ in the $\delta$-net such that $\|\thetav-\tilde\thetav\|_2\leq \delta$) and let $\mathcal{C}_\delta$ denote the mapping of the $\delta$-net under $g_{\thetav}$ described in \prettyref{asmp:diphyp}. This implies that given any $\xv\in \calC$ as in \prettyref{asmp:diphyp}, there is an element in $\calC_\delta$ which is within $\delta$ Euclidean distance of $\xv$.
The choice of $\delta$ that meets our requirements is discussed later. 
For $\tilde \bx\in \calC_\delta$, define $\tilde X,\tilde \Sigma, \Delta\tilde \Sigma$ as in \prettyref{lemma:concent-z}. Then we get that there is a constant $c>0$ independent of $\tilde \bx\in \calC_\delta$ such that
\begin{align}\label{eq:m4}
	&~\PP\qth{\abs{\frac 1{L\sigma_w^2} \sum_{\ell=1}^L \yv_\ell^{\top} \Delta{\tilde \Sigma}\yv_\ell-\Tr(\Delta{\tilde \Sigma}{\Sigma}_o^{-1})}>t}
	\nonumber \\
	&\leq 
	2\exp\Biggl( - c\cdot \min\Biggl\{{Lt^2\over \Tr({\Sigma}_o^{-1}{\Delta \tilde \Sigma}{\Sigma}_o^{-1}{\Delta \tilde \Sigma})},
	{\frac{Lt \cdot  x^4_{\min} (\smin(A))^4}{x_{\max}^2(\smax(A))^4  \|{\xv}_o^2 - {\xvt}^2\|_\infty}}\Biggr\} \Biggr).
\end{align}
Conditioned on the event $\calE_\xi$ in \eqref{eq:E4}, with $m< n/(1+\xi)^2$, the relation \eqref{eq:eigen-singular} and the fact $\|{\xv}_o^2 - {\xvt}^2\|_\infty
\leq x_{\max}^2$ imply that there is a constant $c>0$ depending on $\xi,x_{\min},x_{\max}$ which satisfies
\begin{align*}
&\frac{ x^4_{\min} (\smin(A))^4}{x_{\max}^2(\smax(A))^4  \|{\xv}_o^2 - {\xvt}^2\|_\infty}
\notag\\
&\geq
{x_{\min}^4(\sqrt n-(1+\xi)\sqrt m)^4\over x_{\max}^4(\sqrt n+(1+\xi)\sqrt m)^4}
\geq c.
\end{align*}
In view of the above, we choose 
\begin{align}\label{eq:m18}
	t=\calR\cdot \sqrt{\Tr({\Sigma}_o^{-1}\Delta{\tilde \Sigma}{\Sigma}_o^{-1}\Delta{\tilde \Sigma})}
	+\calR^2,\quad
	\calR=c_2 \sqrt{k\log n\over L},
\end{align}
for a suitably large constant $c_2>0$ to be picked later. Then we have from \eqref{eq:m4} and \eqref{eq:m18}, with a probability at least $1-{e^{-c_3k\log n}}$, for some large constant $c_3>0$ depending on $c_2$
\begin{align}\label{eq:m5}
	&\PP\Biggl[\abs{\frac 1{L\sigma_w^2}\sum_{\ell=1}^L \yv_\ell^{\top}\Delta{\tilde \Sigma}\yv_\ell-\Tr(\Delta{\tilde \Sigma}{\Sigma}_o^{-1})}
	> \calR \sqrt{\Tr({\Sigma}_o^{-1}\Delta{\tilde \Sigma}{\Sigma}_o^{-1}\Delta{\tilde \Sigma})}
		+\calR^2\Biggr]
	\leq 2e^{-c_3k\log n}.
\end{align}
Next, note that from \cite[Chapter 27]{shalev2014understanding}, we have for any set $\calS\subseteq{\reals^k}$ with $\sup_{\btheta\in \calS}\|\btheta\|_\infty< \tau$, the size of any $\delta$-covering set $\calC_\delta$ is bounded as
\begin{align}\label{eq:covering}
	|\mathcal{C}_{\delta}| \leq \pth{\frac{2\tau\sqrt k}{\delta}}^k.
\end{align}
In view of the above, by choosing $\delta=\frac 1{n^8}, \tau=x_{\max}\sqrt{\frac nk}$, we get that it is possible to construct a $\delta$-net $\calC_{\delta}$ of the set $B_{\infty,k} (\mathbf{0}, x_{\max} \sqrt{n \over k})$ of size at most $|\calC_\delta|\leq e^{c_0 k \log n}$ for a constant $c_0>0$. Note that $g_{\thetav}(\cdot)$ is Lipschitz with Lipschitz constant to be 1, and $\tilde{\xv} = g_{\tilde\thetav}(\uv)$.  In view of the above covering number bound, we can pick $c_2>0$ in \eqref{eq:m5} large enough such that $c_3>2c_0$, and we continue \eqref{eq:m5} to get
\begin{align}\label{eq:m13}
	&\PP\Biggl[\abs{\frac 1{L\sigma_w^2}\sum_{\ell=1}^L \yv_\ell^{\top}\Delta{\tilde \Sigma}\yv_\ell-\Tr(\Delta{\tilde \Sigma}{\Sigma}_o^{-1})}
	>\calR  \sqrt{\Tr({\Sigma}_o^{-1}\Delta{\tilde \Sigma}{\Sigma}_o^{-1}\Delta{\tilde \Sigma})}
		+\calR^2,~\forall \tilde \bx: \tilde\thetav \in \calC_\delta\Biggr]
	\leq 2e^{-c_0k\log n}.
\end{align}
Next we pick $\xvt$ as the closest vector in $\calC_\delta$ to $\xvh$, i.e.,
\[
\xvt=\argmin_{\xv\in\calC_\delta}\|\xvh-\xv\|_2.
\]
This implies that $\|\xvh-\xvt\|_2\leq \delta$. To this end, we use \prettyref{lem:delta-error} to get that the event
\begin{align*}
	\calE_2 = 
	&\Biggl\{\abs{\frac 1{L\sigma_w^2}\sum_{\ell=1}^L \yv_\ell^{\top}\Delta{ \Sigma}\yv_\ell-\Tr(\Delta{\Sigma} {\Sigma}_o^{-1})}
	\leq \calR \sqrt{\Tr({\Sigma}_o^{-1}\Delta{\Sigma}{\Sigma}_o^{-1}\Delta{\Sigma})}
		+\calR^2+{\tilde C\over n^3}\Biggr\}
\end{align*}
satisfies
\begin{align}\label{eq:E2}
	\PP\qth{\calE_2}
	\geq 1- O\pth{e^{-c_4k\log n}+e^{-\xi^2\frac m2} +e^{-{Ln\over 8}}}.
\end{align}
\subsubsection{Upper bounding $\Tr({\Sigma}_o^{- 1}\Delta{\Sigma}{\Sigma}_o^{-1}\Delta{\Sigma})$}
Combining \eqref{eq:key:equation}, \eqref{eq:average:likelihood:lower}, and \eqref{eq:upper:lambda:max}, we have conditioned on the event $\calE_\xi$
\begin{align}
	\label{eq:m3}
	&\frac {\Tr({\Sigma}_o^{- 1}\Delta{\Sigma}{\Sigma}_o^{-1}\Delta{\Sigma})}{2(1+\tilde c)^2}
	\leq -\qth{\frac 1{L\sigma_w^2} \sum_{\ell=1}^L \yv_\ell^{\top} \Delta{\Sigma}\yv_\ell-\Tr(\Delta{\Sigma}{\Sigma}_o^{-1})}.
\end{align}
Combining \eqref{eq:m3} and \eqref{eq:E2}, we can see that conditioned on the event $\calE_2\cap \calE_\xi$, we have
\begin{align}
	\label{eq:m6}
	&\frac 1{2(1+\tilde c)^2}\Tr({\Sigma}_o^{- 1}\Delta{\Sigma}{\Sigma}_o^{-1}\Delta{\Sigma})
	\notag\\
	&\leq \calR \sqrt{\Tr({\Sigma}_o^{-1}\Delta{\Sigma}{\Sigma}_o^{-1}\Delta{\Sigma})}
	+\calR^2+{\tilde C\over n^3}.
\end{align}
Define
\begin{align}\label{eq:m7}
	\begin{gathered}
		z=\sqrt{\Tr({\Sigma}_o^{- 1}\Delta{\Sigma}{\Sigma}_o^{-1}\Delta{\Sigma})},
	\quad 
	a=\frac 1{2(1+\tilde c)^2},\\
	b=\calR,
	\quad
	c=\calR^2+{\tilde C\over n^3}.
	\end{gathered}
\end{align}
Then, we can reduce the last inequality to
\begin{align*}
	az^2-bz-c\leq 0,
\end{align*}
which implies,
$$
{b-\sqrt{b^2+4ac}\over 2a} \leq z\leq {b+\sqrt{b^2+4ac}\over 2a},
$$
or in other words, as $z$ in \eqref{eq:m7} is positive, we get
$$
z^2\leq 2\pth{{b^2\over 4a^2}+{b^2+4ac\over 4a^2}}
={b^2\over a^2}+{2c\over a}.
$$
Recalling the definitions from \eqref{eq:m7}, the above inequality transforms into
\begin{align}\label{eq:m21}
	&\Tr({\Sigma}_o^{- 1}\Delta{\Sigma}{\Sigma}_o^{-1}\Delta{\Sigma})
	\notag\\
	&\leq {4(1+\tilde c)^4}\cdot \calR^2 + 8(1+\tilde c)^2(\calR^2+{\tilde C\over n^3})
	\notag\\
	&\leq 4(1+\tilde c)^4\sth{3\calR^2 + {\tilde C\over n^3}}.
\end{align}

\subsubsection{Lower bounding $\Tr({\Sigma}_o^{-1}\Delta{\Sigma}{\Sigma}_o^{-1}\Delta{\Sigma})$ in terms of $\|\texorpdfstring{\xv}{by}_o - \texorpdfstring{\xvh}{by}\|_2$} We will use the following result.

Using Lemma \ref{lemma:vector} and the notations defined in \eqref{eq:proof-notations}, it follows that 
\begin{align}\label{eq:m20}
	&\Tr({\Sigma}_o^{-1}\Delta{\Sigma}{\Sigma}_o^{-1}\Delta{\Sigma})
	\notag\\
	&\geq {x_{\min}^4\lambda_{\min}^2({A}{A}^{\top})\over x_{\max}^8\lambda_{\max}^4({A}{A}^{\top})}
	\|{A}(\hat {X}^2 - {X}_o^2){A}^{\top}\|_{\sf HS}^2.
\end{align}
To obtain a lower bound for $\|{A}(\hat {X}^2 - {X}_o^2){A}^{\top}\|_{\sf HS}^2$ we first write:
\[
\hat{X}^2 = \hat{X}^2 -\tilde{X}^2 + \tilde{X}^2, 
\]
where $\tilde X =\diag(\tilde \bx)$ is chosen from the $\delta$-net with $\delta=\frac 1{n^8}$, such that $\|\hat \bx-\tilde \bx\|_2\leq \delta$. Using 
$$(\|B+C\|_{\sf HS}^2\leq 2\pth{\|B\|_{\sf HS}^2+\|C\|_{\sf HS}^2})$$ 
we get
\begin{align}\label{eq:m17}
	&\|{A}(\hat {X}^2 - {X}_o^2){A}^{\top}\|_{\sf HS}^2
	\geq \frac 12\|{A}(\tilde {X}^2 - {X}_o^2){A}^{\top}\|_{\sf HS}^2 - \|{A}( \tilde{X}_o^2-\hat {X}^2){A}^{\top}\|_{\sf HS}^2.
\end{align}
We bound $\|{A}(\tilde {X}^2 - \hat {X}^2){A}^{\top}\|_{\sf HS}^2$ from above via inequalities on the Hilbert-Schmidt norm and the operator norm, \cite[Chapter IX]{conway2019course}, conditioned on the high probability event $\calE_\xi$ as in \eqref{eq:E4}
\begin{align}\label{eq:m11}
	&\|{A}(\tilde {X}^2 - \hat {X}^2){A}^{\top}\|_{\sf HS}^2
	\notag\\
	&\leq (\sigma_{\max}(A))^4\|\tilde {X}^2 - \hat {X}^2\|_{\sf HS}^2
	\notag\\
	&\leq (2x_{\max})^2n^2\|\tilde {\bx} - \hat {\bx}\|_2^2
	\leq (2x_{\max})^2n^2\delta^2.
\end{align}

Then it remains to find a lower bound for $\|{A}(\tilde {X}^2 - {X}_o^2){A}^{\top}\|_{\sf HS}$ in terms of $\|\xv_o - \xvh\|_2$.

Towards this goal, for $\gamma>0$, define $\Ec_1 (\gamma)$ as the event below 
\begin{align}\label{eq:m22}
	\Ec_1 (\gamma)
	=\|A(\tilde{X}^2-X_o^2)A^{\top}\|^2_{\sf HS}\geq m(m-1) \|\xv_o^2-\xvt^2\|_2^2 - m^2 n \gamma,
\end{align}
Using Lemma \ref{lem:lowerAX2AT}, we show that for an appropriate choice of $\gamma$, this event holds with a high probability.

More precisely, from Lemma \ref{lem:lowerAX2AT},
\begin{align*}
	&~\Prob (\Ec_1^c(\gamma)) \notag\\
	&\overset{(a)}{\leq} 2C e^{k \log \frac{2k}{\delta}}   \exp \left( -c \cdot \min \Big( \frac{\check{\alpha}_{m,n} \gamma^2}{x_{\max}^8}, \frac{\check{\beta}_{m,n} \gamma}{x_{\max}^4} \Big)\right)+ 2 e^{k \log \frac{2k}{\delta} - n/2} \nonumber \\
	&\leq 2mC e^{k \log \frac{2k}{\delta}} \left(  e^{-c \frac {\check{\alpha}_{m,n} \gamma^2}{x_{\max}^8}}+  e^{ -c  \frac{\check{\beta}_{m,n} \gamma}{ x_{\max}^4}}  \right) + 2 me^{k \log \frac{2k}{\delta} - \frac{n}{2}} \nonumber \\
	&=2mC e^{k \log \frac{2k}{\delta}}   e^{ -  \frac{cm^2 \gamma^2}{C^2   x^8_{\max}  (2  + \sqrt{m/n})^4 }} +2C e^{k \log \frac{2k}{\delta}}   e^{ - \frac{cm^2  \gamma}{C x_{\max}^4 (2 + \sqrt{m/n})^2} } \notag\\
	&\quad + 2 e^{k \log \frac{2k}{\delta} - n/3},
\end{align*}
where $\calE_1^c(\gamma)$ denotes the complement of the event $\calE_1(\gamma)$ defined in \eqref{eq:m22}, we define
\begin{eqnarray}
	\check{\alpha}_{m,n} &\triangleq& \frac{m^4n^2 }{C^2  m^2  (2 \sqrt{n} + \sqrt{m})^4 } = \frac{m^2n^2 }{C^2   (2 \sqrt{n} + \sqrt{m})^4 }, \nonumber \\
	\check{\beta}_{m,n}  &\triangleq&  \frac{m^2 n }{C  (2 \sqrt{n} + \sqrt{m})^2},
\end{eqnarray}
and (a) followed by combining \eqref{eq:covering}, a union bound on the choice of $\tilde{\bx}$ and \prettyref{lem:lowerAX2AT}. By setting 
\begin{equation}
	\gamma= 2C \frac{x_{\max}^4 (2+ \sqrt{m/n})^2 }{m \sqrt{c}} \sqrt{k \log \frac{2k}{\delta}}, 
\end{equation}
we have on the event $\calE_1=\calE_1(\gamma)$
\begin{align}\label{eq:lower:axat}
	&\|A (\tilde{X} ^2 - X_o^2 )A^{\top}\|_{\sf HS}^2 
	\notag\\
	&\geq m(m-1) \sum_{i}^n (\tilde{x}_{i}^2 - x_{o,i}^2)^2\nonumber - \tilde{C} mn \sqrt{k \log \frac{2k}{\delta}} \nonumber \\
	&= m(m-1) \sum_{i}^n (\tilde{x}_{i} - x_{o,i})^2 (\tilde{x}_{i} + x_{o,i})^2 - \tilde{C} mn \sqrt{k \log \frac{2k}{\delta}} \nonumber \\
	&\geq 4m(m-1) x_{\min}^2 \sum_{i}^n (\tilde{x}_{i} - x_{o,i})^2 \nonumber - \tilde{C} mn \sqrt{k \log \frac{2k}{\delta}} \nonumber \\
	&= 4m(m-1) x_{\min}^2 \|\xvt- \xv_o\|_2^2  - \tilde{C} mn \sqrt{k \log \frac{2k}{\delta}},
\end{align}
with probability
\begin{align}
	\label{eq:E1}
	\Prob (\Ec_1^c) \leq O(me^{-k \log \frac{k}{\delta}}   +  e^{k \log \frac{k}{\delta}-\frac{n}{3}}).
\end{align}
In the above equations $\tilde{C}$ is a constant that does not depend on $m,n$ or $\delta$. Furthermore, in the last display we have assumed that $m$ is large enough (and hence $\gamma$ is small enough) to make the inequality $\frac{m^2 \gamma^2}{C^2   x^8_{\max}  (2  + \sqrt{m/n})^4 }< \frac{m^2  \gamma}{C x_{\max}^4 (2 + \sqrt{m/n})^2}$ true. Simplifying \eqref{eq:lower:axat}, with 
$$
\|\hat\bx-\bx_o\|_2\leq \|\tilde \bx- \bx_o\|_2 + \|\hat\bx-\tilde \bx\|_2
\leq \|\tilde\bx- \bx_o\|_2+\delta,
$$
we use $(b+c)^2\leq 2(b^2+c^2)$ with $b=\|\tilde\bx- \bx_o\|_2,c=\delta$ to get
\begin{align}\label{eq:lower:axat_2}
	&\|A (\tilde{X} ^2 - X_o^2 )A^{\top}\|_{\sf HS}^2
	\notag\\
	&\geq 
	4m(m-1) x_{\min}^2 \|\xvt- \xv_o\|_2^2  - \tilde{C} mn \sqrt{k \log \frac{2k}{\delta}}
	\nonumber\\
	&\geq 4m(m-1) x_{\min}^2 \pth{\frac 12\|\hat \bx- \xv_o\|_2^2-\delta^2}  - \tilde{C} mn \sqrt{k \log \frac{2k}{\delta}}.
\end{align}
In view of \eqref{eq:m17}, we combine \eqref{eq:m20},\eqref{eq:m11},\eqref{eq:lower:axat_2} to get
\begin{align*}
	&~{x_{\max}^8\lambda_{\max}^4({A}{A}^{\top})\over x_{\min}^4\lambda_{\min}^2({A}{A}^{\top})}\Tr({\Sigma}_o^{-1}\Delta{\Sigma}{\Sigma}_o^{-1}\Delta{\Sigma})
	\notag\\
	&\geq \frac 12\|{A}(\tilde {X}^2 - {X}_o^2){A}^{\top}\|_{\sf HS}^2-(2x_{\max})^2n^2\delta^2
	\nonumber\\
	&\geq
	\frac {4m(m-1) x_{\min}^2}2\pth{\frac 12\|\hat \bx- \xv_o\|_2^2-\delta^2}  
	\quad - \tilde{C} mn \sqrt{k \log \frac{2k}{\delta}}-(2x_{\max})^2n^2\delta^2.
\end{align*}

\subsubsection{Combining the results} 
In view of \eqref{eq:m21} and the last display, on the event $\calE_\xi$ as in \eqref{eq:E4} we get
\begin{align*}
	&\frac {4m(m-1) x_{\min}^2}2\pth{\frac 12\|\hat \bx- \xv_o\|_2^2-\delta^2} 
	\notag\\
	&- \tilde{C} mn \sqrt{k \log \frac{2k}{\delta}}-(2x_{\max})^2n^2\delta^2
	\nonumber\\
	&\leq \tilde C_2{ x_{\max}^8\lambda_{\max}^4({A}{A}^{\top})\over x_{\min}^4\lambda_{\min}^2({A}{A}^{\top})}
	\cdot \sth{\calR^2 + {1\over n^3}}
	\notag\\
	& \leq \tilde C_3n^2\pth{{k\log n\over L}  + {1\over n^3}},
\end{align*}
for constants $\tilde C,\tilde C_2,\tilde C_3 >0$ depending on $x_{\min},x_{\max}$. 
Simplifying with $\delta=\frac 1{n^8}$, and \eqref{eq:E1},\eqref{eq:E2},\eqref{eq:E4} we get  
$$
\frac 1n \|\hat \bx- \bx_o\|_2^2 
\leq \tilde C_4\pth{{n\over m^2}\cdot {k\log n\over L}+{\sqrt{k \log{n}}\over m}},
$$
for a constant $\tilde C_4>0$ depending on $x_{\min},x_{\max}$, on the event $\calE = \calE_1\cap\calE_2\cap\calE_\xi$ with $\PP[\calE]\geq 1- C_2\pth{e^{-C_3{m}} + e^{-C_3{Ln}} + me^{-C_3k \log n}+ e^{C_4k \log n-C_3{n}}}$ for some constants $C_2,C_3,C_4>0$. This completes our proof of \prettyref{thm:maintheorem}.


\section{Conclusion}
We have explored the theoretical and algorithmic aspects of the problem of signal recovery from multiple sets of measurements, termed as looks, amidst the presence of speckle noise. We established an upper bound on the MSE of such imaging systems, effectively capturing the MSE's dependence on the number of measurements, image complexity, and number of looks. Drawing inspiration from our theoretical framework, we introduce the Bagged Deep Image Prior (Bagged-DIP) projected gradient descent (PGD) algorithm. Through extensive experimentation, we demonstrate that our algorithm attains state-of-the-art performance.

\section*{Acknowledgements}
X.C., S.Jalali and A.M. were supported in part by ONR award no. N00014-23-1-2371. S.Jalali was supported in part by NSF CCF-2237538. C.A.M. was supported in part by SAAB, Inc., AFOSR Young Investigator Program Award no. FA9550-22-1-0208, and ONR award no. N00014-23-1-2752.

\bibliographystyle{IEEEtran}
\bibliography{myrefs}

\appendices

\section{Technical results}
\label{app:technical}

We present a few lemmas that are used in the proof of \prettyref{thm:maintheorem}.

\subsection{Proof of \prettyref{lemma:concent-z}}
	Define matrix $B=X_oA^{\top}{\Delta \tilde \Sigma} AX_o\in \mathbb{R}^{n\times n}$ and $\tilde{B}\in \mathbb{R}^{Ln\times Ln}$ as
	\[
	\tilde{B} = \begin{bmatrix}
		B & 0 & \cdots & 0 \\
		0 & B & \cdots & 0 \\
		0 & 0 & \cdots & 0 \\
		0 & 0 & \cdots & B \\
	\end{bmatrix}.
	\]
	Furthermore, for $\bw_i\simiid \calN(\mathbf{0},\sigma_w^2\bI_{n})$, define
	\[
	{\bW}^{\top} = [\wv_1^{\top}, \wv_2^{\top}, \ldots, \wv_L^{\top}]\in \reals^{Ln}. 
	\]
	Note that for any fixed $\Delta \tilde \Sigma$ we use $\Sigma_o^{-1}=AX_o^2A^\top$ to get
	\begin{align*}
		&\EE\qth{\bW^{\top} \tilde B\bW}
		=L\EE[\bw_1^{\top} B \bw_1]
		=L\EE[\Tr(B\bw_1\bw_1^{\top})]
		\notag\\
		&\stepa{=}L\sigma_w^2\Tr({\Delta \tilde \Sigma} A X_o^2 A^{\top})
		=L\sigma_w^2\Tr({\Delta \tilde \Sigma}\Sigma_o^{-1}).
	\end{align*}
	where (a) follows using $\Tr(AB)=\Tr(BA)$ for two matrices $A,B$. To get a concentration bound on $\bW^\top \tilde B \bW$, we use the Hanson-Wright inequality given below.
	\begin{lemma}[Hanson-Wright inequality \cite{RV13}] \label{lem:HW-ineq}
		Define $\|X\|_{\psi_2} = \inf\{t>0 : \mathbb{E} (\exp (X^2/t^2)) \leq 2\} 
		$. Let $\Xv= (X_1,...,X_n)$ be a random vector with independent components satisfying $\E[X_i]=0$ and $\|X_i\|_{\Psi_2}\leq K$. Let A be an $n\times n$ matrix. Then, there is a constant $c>0$ such that 
		\begin{align*}
			&\Prob \Big(|\Xv^{\top} A\Xv-\E[\Xv^{\top} A\Xv]|>t \Big) 
			\notag\\
			&\leq 2\exp\left(-c\min\left({t^2\over K^4\|A\|_{\sf HS}^2 },{t \over K^2\|A\|_2 }\right)\right),
			\quad 
			t>0.
		\end{align*}
	\end{lemma}
	We substitute $A$ with $\tilde B$ and $\bX$ with $\frac 1{\sigma_w}\bW$ in the above result. Note that, $K$ is at most a constant as each coordinate of $\frac 1{\sigma_w}\bW$ has the standard normal distribution. Then, we get for a constant $c>0$
	\begin{align}
		\label{eq:m16}
		&\Prob\qth{\abs{{1\over L \sigma_w^2}\bW^{\top}\tilde{B}\bW- {\rm Tr}({\Delta \tilde \Sigma}\Sigma_o^{-1})}>t}
		\notag\\
		&\leq 2\exp\Bigg(-c \cdot \min\Biggl\{{L^2 t^2\over \|\tilde{B}\|_{\sf HS}^2 },{L t \over \|\tilde{B}\|_2 }\Biggr\}\Bigg).
	\end{align}
	Next, note that
	\begin{align}\label{eq:hs-L2-1}
		\|\tilde{B}\|_{\sf HS}^2&=L {\rm Tr}(B^2)
		=L\Tr(X_oA^{\top} {\Delta \tilde \Sigma} A X_o^2A^{\top} {\Delta \tilde \Sigma A} X_o)
		\notag\\
		&=L\Tr(A X_o^2A^{\top}{\Delta \tilde \Sigma} A X_o^2A^{\top} {\Delta \tilde \Sigma}),
	\end{align}
	and 
	\begin{align}\label{eq:hs-L2-2}
		&~\|\tilde{B}\|_2
		= \|B\|_2 = \|{X}_o{A}^{\top}{\Delta \tilde \Sigma}{A}{X}_o\|_2
		\nonumber\\
		&\leq 
		x_{\max}^2\sigma_{\max}(A)\sigma_{\max}(A^{\top})
		\sigma_{\max}(\Delta\tilde \Sigma)
		\notag\\
		&\stepa{\leq } \frac{x_{\max}^2(\sigma_{\max}(A))^2\sigma_{\max} (A {X}_o^2 A^{\top}-A \tilde{X}^2 A^{\top} )}{\sigma_{\min} (A\hat{X}^2 A^{\top}) \sigma_{\min} (A\tilde{X}^2 A^{\top}) }
		\nonumber\\
		&
		\leq \frac{x_{\max}^2(\sigma_{\max}(A))^2\lambda_{\max}(AA^{\top})  \|{\xv}_o^2 - {\xvt}^2\|_\infty}{x^4_{\min} \lambda_{\min}^2 (AA^{\top})}
		\notag\\
		&\stepb{\leq} 
		\frac{x_{\max}^2(\sigma_{\max}(A))^4  \|{\xv}_o^2 - {\xvt}^2\|_\infty}{x^4_{\min}  (\smin(A))^4},
	\end{align}
	where the last inequality followed from \eqref{eq:eigen-singular} and (a) followed using the result below. 
	\begin{lemma}\label{lem:boundeigenvalues}
		Let $B$ and $C$ denote two $n \times n$ symmetric and invertible matrices. Then, if $\lambda_i$ represents the $i^{\rm th}$ eigenvalue of $B^{-1}- C^{-1}$, we have $|\lambda_i| \in [-\frac{\sigma_{\max} (B-C)}{\sigma_{\min}(B) \sigma_{\min}(C) }, \frac{\sigma_{\max} (B-C)}{\sigma_{\min}(B) \sigma_{\min}(C) }]$.
	\end{lemma}
	Substituting \eqref{eq:hs-L2-1},\eqref{eq:hs-L2-2} in \eqref{eq:m16} we conclude the proof. We end this section with a proof of \prettyref{lem:boundeigenvalues}.
	\begin{proof}[Proof of \prettyref{lem:boundeigenvalues}]
		Suppose $\lambda_i$ is the $i^{\rm th}$ eigenvalue of $B^{-1}-C^{-1}$. Then, there exists a unit-norm vector $\vv \in \mathbb{R}^n$ such that 
		$
			(B^{-1}-C^{-1}) \vv = \lambda_i \vv. 
		$
		Multiplying both sides by $B$, we have
		$
			(I-BC^{-1}) \vv = \lambda_i B \vv. 
		$
		Define $\uv = C^{-1}\vv$. Then, we have
		$(C-B) \uv = \lambda_i BC \uv,$ or equivalently
		$
		\lambda_i  \uv = (BC)^{-1} (C-B) \uv. 
		$
		Hence,
		$
		|\lambda_i| \leq \frac{\sigma_{\max}(C-B)}{\sigma_{\min} (B) \sigma_{\min} (C)}. 
		$
	\end{proof}

\subsection{Proof of \prettyref{lem:delta-error}}

	\begin{enumerate}[label=(\alph*.)]
		\item Rearranging the terms in the expression 
		$h(\hat \Sigma)-h(\tilde \Sigma)$, we get that it suffices to prove
		$$
		\abs{\frac 1{L\sigma_w^2} \sum_{\ell=1}^L \yv_\ell^{\top} {(\tilde \Sigma-\hat\Sigma)}\yv_\ell}
		+\abs{\sum_{i=1}^m\lambda_i({\Sigma}_o^{-\frac 12}(\hat \Sigma-\tilde \Sigma){\Sigma}_o^{-\frac 12})}
		\leq 
		\tilde C n\delta.
		$$
		Our proof uses the following result. A proof is provided at the end of this section.
		\begin{lemma}\label{lem:upper}
			Assume that both $\tilde \Sigma = (A\tilde{X}^2A^{\top})^{-1}$ and $\hat\Sigma = (A\hat{X}^2A^{\top})^{-1}$ exist. Then,
			\begin{equation*}
				|\lambda_i ({\Sigma}_o^{-{1\over 2}}(\hat \Sigma - \tilde\Sigma) {\Sigma}_o^{-{1\over 2}}) | \in [0, \frac{x_{\max}^2\lambda^2_{\max} (A A^{\top}) \| {\xvh}^2 - {\xvt}^2\|_\infty}{x_{\min}^4 \lambda^2_{\min} (AA^{\top})} ].
			\end{equation*}
			Furthermore, we have
			\begin{align*}
				&\abs{\frac{1}{L\sigma_w^2}\sum_{\ell =1}^L \yv_\ell^{\top} (\hat \Sigma - \tilde\Sigma) \yv_{\ell}} 
				\notag\\
				&\leq  \frac{x_{\max}^2 \lambda^2_{\max}(AA^{\top})  \|{\xvh}^2 - {\xvt}^2\|_\infty}{x^4_{\min} \lambda_{\min}^2 (AA^{\top})} \cdot \frac{1}{L\sigma_w^2} \sum_{\ell=1}^L \wv_{\ell}^{\top} \wv_{\ell}.  
			\end{align*}
		\end{lemma}
		 In view of the above result, we first bound the following three terms from above:
		\begin{itemize}
			\item We first bound $\frac{1}{L \sigma_w^2} \sum_{\ell} \wv_{\ell}^{\top} \wv_{\ell}$. To this end, we use the following result.
			\begin{lemma}[Concentration of $\chi^2$ \cite{jalali2014minimum}]\label{lem:conc:chisq}
				Let $Z_1, Z_2, \ldots, Z_n$ denote a sequence of independent $\Nc(0,1)$ random variables. Then, for any $t\in(0,1)$, we have
				\[
				\Prob (\sum_{i=1}^n Z_i^2 \leq n (1-t)) \leq e^{\frac{n}{2} (t + \log (1- t)) }. 
				\]
				Also, for any $t>0$,
				\begin{align}
					\label{eq:m2}
				\Prob (\sum_{i=1}^n Z_i^2 \geq n (1+t)) \leq e^{-\frac{n}{2} (t -\log (1+ t)) }. 
				\end{align}
			\end{lemma}
			In view of the above, as $\sth{\frac {w_\ell}{\sigma_w}}_{\ell=1}^L$ are independent $N(0,1)$ random variables,  we get
			\begin{align*}
			\Prob\qth{ \frac{1}{L \sigma_w^2} \sum_{\ell=1}^{L} \wv_{\ell}^{\top} \wv_{\ell} \leq 2n }\geq 1-e^{-\frac{Ln}{8}}.
			\end{align*}
			
			\item Next, we bound $\frac{\lambda^2_{\max}(AA^{\top})}{ \lambda_{\min}^2 (AA^{\top})} $. Consider the event $\calE_\xi$ described in \eqref{eq:E4}. In view of \prettyref{lem:singvalues} we get $\PP[\calE_\xi]\geq 1-2e^{-\xi^2\frac m2}$. Hence, conditioned on $\calE_\xi$ it is straightforward to see that
			\begin{align}\label{eq:E_xi-bound}
				&\frac{ \lambda^2_{\max}(AA^{\top})  }{ \lambda_{\min}^2 (AA^{\top})}  \leq \frac{(\sqrt{n} + (1+\xi)\sqrt{m})^2}{(\sqrt{n} - (1+\xi)\sqrt{m})^2}, 
			\end{align}
			
			\item Finally, we will bound $\|{\xvh}^2 - {\xvt}^2\|_\infty$. In view of the Lipschitz property of $g_{\tilde{\thetav}} (\uv)$ provided in \prettyref{asmp:diphyp}, we get that
			\begin{align*}
			&\|\xvh^2- \xvt^2\|_\infty \leq 2 x_{\max} \|\xvh- \xvt\|_\infty \leq 2 x_{\max}	\|\xvt-\hat{\xv}\|_2 \notag\\
			&\leq 2x_{\max}\|\tilde \btheta - \hat\btheta\|_2\leq 2x_{\max}\delta 
			\end{align*}
		\end{itemize}
		Summarizing the discussions above, we can conclude from \prettyref{lem:upper}
		\begin{align}\label{eq:m9}
			&~\PP\qth{\abs{\frac{1}{L\sigma_w^2}\sum_{\ell =1}^L \yv_\ell^{\top} ( \hat\Sigma - \tilde \Sigma) \yv_{\ell}}
			\leq C_1 n\delta}
			\geq 1-2e^{-\xi^2\frac m2}-e^{-{Ln\over 8}},
		\end{align}
		for a constant $ C_1>0$.
		To complete the proof we exhibit the following bound on $\abs{\lambda_{\max}({\Sigma}_o^{-\frac 12}(\hat \Sigma-\tilde\Sigma){\Sigma}_o^{-\frac 12}}$
		\begin{align*}
			&\abs{\lambda_{\max}(\Sigma_o^{-\frac 12}(\hat \Sigma-\tilde\Sigma)\Sigma_o^{-\frac 12})}
			\leq \frac{x_{\max}^2 \lambda^2_{\max}(AA^{\top})  \|{\xvh}^2 - {\xvt}^2\|_\infty}{x^4_{\min} \lambda_{\min}^2 (AA^{\top})}
			\leq C_2\delta,
		\end{align*}
		for a constant $C_2>0$ that may depend on $\xi,x_{\min},x_{\max}$, conditioned on the event $\calE_\xi$ and we use \eqref{eq:E_xi-bound}. Hence, we get
		\begin{align}\label{eq:m10}
			\PP\qth{\abs{\sum_{i=1}^m\lambda_i({\Sigma}_o^{-\frac 12}(\hat \Sigma-\tilde \Sigma){\Sigma}_o^{-\frac 12})}
			\leq 
			\tilde C_2 n\delta}
			\geq 1-e^{-\xi^2\frac m2}.
		\end{align}
		Our proof then follows by combining \eqref{eq:m9}, \eqref{eq:m10}. We end our argument with a proof of \prettyref{lem:upper}.
		\begin{proof}[Proof of \prettyref{lem:upper}]
			To prove the first inequality, we note that
			\begin{align*}
				&|\lambda_i({\Sigma}_o^{-{1\over 2}}(\hat \Sigma - \tilde\Sigma) {\Sigma}_o^{-{1\over 2}})|
				\leq   \frac{\abs{\sigma_{\max} (\hat \Sigma - \tilde\Sigma)}}{\sigma_{\min}({\Sigma_o})}
				\notag\\
				&= \frac{\abs{\sigma_{\max} ((A\hat{X}^2 A^{\top})^{-1} -(A\tilde{X}^2 A^{\top})^{-1} )}}{\sigma_{\min}({\Sigma_o})} \nonumber \\
				&\overset{(a)}{\leq} \frac{\abs{\sigma_{\max} (A\hat{X}^2 A^{\top}-A\tilde{X}^2 A^{\top} )}}{\sigma_{\min}({\Sigma}_o) \sigma_{\min} (A\hat{X}^2 A^{\top}) \sigma_{\min} (A\tilde{X}^2 A^{\top}) }   \nonumber \\
				&= \frac{\sigma_{\max}(A {X}_o^2 A^{\top})  \abs{\sigma_{\max} (A(\hat{X}^2 - \tilde{X}^2 )A^{\top}) }}{\sigma_{\min} (A\hat{X}^2 A^{\top}) \sigma_{\min} (A\tilde{X}^2 A^{\top}) }
				\notag\\
				&\leq \frac{x_{\max}^2 \lambda^2_{\max}(AA^{\top})  \|{\xvh}^2 - {\xvt}^2\|_\infty}{x^4_{\min} \lambda_{\min}^2 (AA^{\top})}.
			\end{align*}
			where we obtain inequality (a) by using \prettyref{lem:boundeigenvalues}. To prove the second inequality, note that 
			\begin{align*}
				\abs{\frac{1}{L\sigma_w^2}\sum_{\ell =1}^L \yv_\ell^{\top} (\hat \Sigma - \tilde\Sigma) \yv_{\ell}}
				&\leq  \abs{\sigma_{\max } ( (A \hat{X}^2A^{\top})^{-1} - (A \tilde{X}^2A^{\top})^{-1} )} \frac{1}{L \sigma_w^2} \sum_{\ell=1}^L \yv_{\ell}^{\top} \yv_{\ell} \nonumber \\
				&\stepb{\leq}  \frac{ \lambda_{\max}(AA^{\top})  \|{\xvh}^2 - {\xvt}^2\|_\infty}{x^4_{\min} \lambda_{\min}^2 (AA^{\top})} \frac{1}{L \sigma_w^2} \sum_{\ell=1}^L \yv_{\ell}^{\top} \yv_{\ell} 
				\notag\\
				&\leq \frac{x_{\max}^2 \lambda^2_{\max}(AA^{\top})  \|{\xvh}^2 - {\xvt}^2\|_\infty}{L \sigma_w^2 x^4_{\min} \lambda_{\min}^2 (AA^{\top})} \sum_{\ell=1}^L \wv_{\ell}^{\top} \wv_{\ell},
			\end{align*}
			where inequality (b) followed from a similar analysis as in the proof of the first part. 
		\end{proof}
		
		\item To prove the second result, in view of \eqref{eq:m9}, as $\lambda_{\max}(AB)=\lambda_{\max}(BA)$, we get on the event $\calE_\xi$
		\begin{align}\label{eq:m15}
			&~\Tr({\Sigma}_o^{-1}\Delta{\tilde \Sigma}{\Sigma}_o^{-1}\Delta{\tilde \Sigma})-\Tr({\Sigma}_o^{-1}\Delta{ \Sigma}{\Sigma}_o^{-1}\Delta{ \Sigma})
			\notag\\
			&=\Tr(({\Sigma}_o^{-1}\Delta{\tilde \Sigma})^2)-\Tr(({\Sigma}_o^{-1}\Delta{ \Sigma})^2)
			\nonumber\\
			&=\Tr(({\Sigma}_o^{-1}(\hat \Sigma-\tilde\Sigma))({\Sigma}_o^{-1}(\Delta{\tilde \Sigma}+\Delta{ \Sigma})))
			\nonumber\\
			&\leq n |\lambda_{\max}({\Sigma}_o^{-1}(\hat \Sigma-\tilde\Sigma))|\cdot |\lambda_{\max}({\Sigma}_o^{-1}(\Delta{\tilde \Sigma}+\Delta{ \Sigma}))|
			\leq \tilde C_2 n\delta
			\cdot 
			|\lambda_{\max}({\Sigma}_o^{-1}(\Delta{\tilde \Sigma}+\Delta{ \Sigma}))|.
		\end{align}
		To get an upper bound on $|\lambda_{\max}({\Sigma}_o^{-1}(\Delta{\tilde \Sigma}+\Delta{ \Sigma}))|$, we note that
		{ \begin{align*}
			&~|\lambda_{\max}({\Sigma}_o^{-1}(\Delta{\tilde \Sigma}+\Delta{ \Sigma}))|
			\leq   \frac{\sigma_{\max} (\Delta\tilde \Sigma+\Delta \Sigma)}{\sigma_{\min}({\Sigma_o})} 
			\nonumber \\
			&=\frac{\abs{\sigma_{\max} ((A\hat{X}^2 A^{\top})^{-1} +(A\tilde{X}^2 A^{\top})^{-1} -2(A {X}_o^2 A^{\top})^{-1})}}{\sigma_{\min}({\Sigma_o})} \nonumber \\
			&
			\leq \frac{\abs{\sigma_{\max} ((A\hat{X}^2 A^{\top})^{-1})}+\abs{\sigma_{\max}((A\tilde{X}^2 A^{\top})^{-1})}}{\sigma_{\min}({\Sigma_o})} + \frac{2\abs{\sigma_{\max}((A {X}_o^2 A^{\top})^{-1})}}{{\sigma_{\min}({\Sigma_o})}} \nonumber \\
			&= \sigma_{\max}(A {X}_o^2 A^{\top})
			\Biggl(\frac 1{\sigma_{\min}(A\hat{X}^2 A^{\top})}+\frac 1{\sigma_{\min}(A\tilde{X}^2 A^{\top})}
			+\frac 2{\sigma_{\min}(A {X}_o^2 A^{\top})}\Biggr)
			& \leq \frac{4x_{\max}^2 \lambda_{\max}(AA^{\top})}{x^2_{\min} \lambda_{\min} (AA^{\top})}.
		\end{align*}}
		
		Hence, we continue \eqref{eq:m15} to get
		\begin{align*}
			&\Tr({\Sigma}_o^{-1}\Delta{\tilde \Sigma}{\Sigma}_o^{-1}\Delta{\tilde \Sigma})-\Tr({\Sigma}_o^{-1}\Delta{ \Sigma}{\Sigma}_o^{-1}\Delta{ \Sigma})
			\leq \frac{4\tilde C_2 n \delta x_{\max}^2 \lambda_{\max}(AA^{\top})  }{x^2_{\min} \lambda_{\min} (AA^{\top})}
			\leq \tilde C_3 n\delta.
		\end{align*}
		where the last inequality holds on the event $\calE_\xi$ as in \eqref{eq:E4} for a suitable constant $\tilde C_3$. Next we note the fact that for $a,b,c\geq 0$, we have $a-b<c$ means $\sqrt a - \sqrt b \leq \sqrt c$ (as otherwise $\sqrt a - \sqrt b > \sqrt c$ will mean $a>b+c+2\sqrt{bc}$, leading to a contradiction). Hence we have on event $\calE_\xi$
		\begin{align*}
			&\sqrt{\Tr({\Sigma}_o^{-1}\Delta{\tilde \Sigma}{\Sigma}_o^{-1}\Delta{\tilde \Sigma})}-\sqrt{\Tr({\Sigma}_o^{-1}\Delta{ \Sigma}{\Sigma}_o^{-1}\Delta{ \Sigma})}
			\leq \sqrt{\tilde C_3 n\delta}.
		\end{align*}
	\end{enumerate}
	This completes the proof.
\subsection{Proof of \prettyref{lem:lowerAX2AT}}

{	The proof closely follows that of  \cite[Lemma 4 and 5]{zhou2024correction} with modifications to establish our results, which we provide here. 
	Let $\av_{i}^\top $ denote the $i^{\rm th}$ row of matrix $A$. We have
	\begin{align}\label{eq:concent:imp:usedecoup}
		\begin{split}
			\|A DA^\top \|^2_{\rm HS} 
			=\sum_{i=1}^m \sum_{j=1}^m |\av_{i}^\top  D \av_{j}|^2
			\geq \sum_{i \ne j} |\av_{i}^\top  D \av_{j}|^2.
		\end{split} 
	\end{align}
	Hence, we get the inequality
	\begin{align}\label{eq:m19}
		&~\PP \qth{\|A D A^\top \|^2_{\rm HS} < m(m-1) \|\bd\|_2^2 - t}
		\notag\\
		&\le \PP\bigg[  \sum_{i \ne j} |\av_{i}^\top  D \av_{j}|^2 -m(m-1) \sum_{i=1}^n d_i^2< -t \bigg].
	\end{align}
	In view of the above, we only bound the probability on the right hand side. This is established by showing that $\sum_{i \ne j} |\av_{i}^\top  D \av_{j}|^2$ concentrates around $m(m-1) \sum_{i=1}^n d_i^2$ with a high probability. To this end, first consider the following decoupling result.
	\begin{lemma}{\cite[Theorem 3.4.1.]{de2012decoupling}}\label{lmm: abstract decoupling}
		Let $X_1, X_2, \ldots, X_n$ denote random variables with values in measurable space $(S, \mathcal{S})$. Let $(\widetilde{X}_1, \widetilde{X}_2, \ldots, \widetilde{X}_n)$ denote an independent copy of $X_1, X_2, \ldots, X_n$. For $ i \neq j$ let $h_{i,j} : S^2 \rightarrow \mathbb{R}$. Then, there exists a constant $C$ such that for every $t >0$ we have
		\[
		\Prob \qth{\Big|\sum_{i \neq j} h_{i,j} (X_i, X_j) \Big | > t} \leq C \Prob \qth{C \Big|\sum_{i \neq j} h_{i,j} (X_i, \widetilde{X}_j) \Big | > t} . 
		\] 
	\end{lemma}
	By the above result, there exists a constant $C>0$ such that 
	\begin{align}\label{eq:using the abstract decoupling on off diagonal part}
		&~\PP \Bigg(\bigg|  \sum_{i \ne j} |\av_{i}^\top  D \av_{j}|^2 -m(m-1) \sum_{i=1}^n d_i^2\bigg|> t \Bigg)\\ 
			&\leq C \PP \Bigg(C \bigg|  \sum_{i \ne j} |\av_{i}^\top  D \tilde{\av}_{j}|^2 -m(m-1) \sum_{i=1}^n d_i^2\bigg|> t \Bigg)  \\
			&= C  \PP \Bigg(C \bigg|  \sum_{i=1}^m  \av_{i}^\top  D \bigg( \sum_{i \ne j=1}^m \tilde{\av}_{j}  \tilde{\av}_{j}^\top \bigg) D \av_{i} 
			 - m(m-1) \sum_{i=1}^n d_i^2\bigg|> t \Bigg), 
	\end{align}
	where $\tilde \av_{j}$'s denote the independent copies of $\av_{j}$'s for $1\le j \le m$. Define $\tilde{A}$ as the $m\times n$ Gaussian matrix whose rows are $\tilde{\av}_{1}^\top , \ldots, \tilde{\av}_{m}^\top $, i.e.,
	\begin{align*}
		\widetilde A:=\begin{bmatrix}
			\tilde \av_{1}^\top \\ \vdots \\ 
			\tilde \av_{m}^\top
		\end{bmatrix}
	\end{align*}
	
	Also, let $\tilde{A}_{\backslash i}$ denote the matrix that is constructed by removing the $i^{\rm th}$ row of $\tilde{A}$. Define
	\begin{align*}
	&F :=\left[\begin{array}{cccc} D \tilde{A}_{\backslash 1}^\top  \tilde{A}_{\backslash 1} D&0&\ldots&0 \\ 0 & D\tilde{A}_{\backslash 2}^\top  \tilde{A}_{\backslash 2}D&\ldots&0 \\ 0 &0&\ldots& D\tilde{A}_{\backslash m}^\top  {\tilde{A}}_{\backslash m}D \end{array} \right],
	\\
	&\vv:= [\av_{1}^\top , \ldots, \av_{m}^\top ].
	\end{align*}
	Let $\PP_{{\tilde A}}(\cdot):=\PP\qth{\cdot \mid \tilde A}$. Note that for any event $\Ec$, by definition $\PP_{{\tilde A}}(\Ec)=\E[\mathbf{1}_{\Ec}\mid {\tilde A}]$. 
	It follows that 
	\begin{align}\label{eq:off diagonal part 2}
		\begin{split}
			&  ~\PP \Bigg(C \bigg| \sum_{i=1}^m  \av_{i}^\top  D \bigg(\sum_{i \ne j=1}^m \tilde{\av}_{j}  \tilde{\av}_{j}^\top \bigg) D \av_{i} -m(m-1) \sum_{i=1}^n d_i^2\bigg|> t \Bigg)\\
			& = \mathbb{E}\qth{ \PP_{{\tilde A}}\pth{C\abs{\vv^\top  F \vv -\E\qth{\vv^\top  F \vv  | {\tilde{A}} }}\ge t/2}} 
			\notag\\
			&
			\quad + \PP \pth{C\abs{\E\qth{\vv^\top  F \vv  | {\tilde{A}} } - m(m-1)\sum_{i=1}^n d_i^2}\ge t/2}.
		\end{split}
	\end{align}
	Note that from \prettyref{lem:HW-ineq} (with fixed ${\tilde A}$) we have
	\begin{equation}\label{eq:f1}
		\begin{split}
			&\PP_{\tilde A}\pth{C\abs{\vv^\top  F \vv -\E\qth{\vv^\top  F \vv  | {\tilde{A}} }}\ge t/2} \\
		&\le  2\exp \left( -c \min \Bigg( \frac{t^2}{4C^2 K^4\|F\|_{\rm HS}^2 }, \frac{t}{2C K^2\|F\|_2 }\Bigg) \right).
		\end{split}
	\end{equation}
	Define the event $\widetilde \Ec_{\rm maxsing}:= \bigcap_{i=1}^m \cb{\smax(\tilde{A}_{\backslash i})\le \sqrt{m}+2\sqrt{n}}$, and define $E_{\max} = (\sqrt{m}+ 2\sqrt{n})^2$. By \prettyref{lem:singvalues}, we have $\PP(\Ec_{\rm maxsing})\ge 1-me^{-\frac n2}$.
	Restricted to the event $\widetilde \Ec_{\rm maxsing}$, we have
	\begin{align}\label{eq:F2:bound}
		\|F\|_2=\max_{1\le i\le m}\smax\pth{D\tilde{A}_{\backslash i}^\top  \tilde{A}_{\backslash i}D}
		\le \emax\nm{\bd}_{\infty}^2.
	\end{align}
	and using the definition of the HS-norm $\nmhs{F}^2=\Tr(F^\top F)$, we get
	\begin{align}\label{eq:m14}
		&\nmhs{F}^2
		=   \sum_{i=1}^m \trbr{D \tilde{A}_{\backslash i}^\top  \tilde{A}_{\backslash i} DD \tilde{A}_{\backslash i}^\top  \tilde{A}_{\backslash i} D}
		\overset{(a)}{\le}  (m-1)\sum_{i=1}^m\smax(( D\tilde{A}_{\backslash i}^\top  \tilde{A}_{\backslash i} D))^2 
		\le m^2\qth{\emax}^2
		\|\bd\|_{\infty}^4, 
	\end{align}
	where the last inequality followed from \eqref{eq:F2:bound}
	and the inequality (a) followed as the number of nonzero singular values of $D \tilde{A}_{\backslash i}^\top  \tilde{A}_{\backslash i} D$ is equal to its rank, which is at most $(m-1)$ as $\tilde A_{\backslash i}$ has $m-1$ rows, and the result below.
	\begin{lemma}\cite[Theorem 7.4.1.1]{horn2012matrix} \label{lmm:vonnueman} 
		Let $A,B \in \mathbb{R}^{n \times n}$ denote two matrices with singular values $\sigma^{A}_i$ and $\sigma^B_i$. Then
		\(
		{\rm Tr}(AB)  \leq \sum_i \sigma^{A}_i \sigma_i^B.
		\)
	\end{lemma}
	Combining, \eqref{eq:f1}, \eqref{eq:F2:bound}, and \eqref{eq:m14} we have
	\begin{align}\label{eq:off_diag:2}
		\begin{split}
			&\mathbb{E}\qth{ \PP_{{\tilde A}}\pth{C\abs{\vv^\top  F \vv -\E\qth{\vv^\top  F \vv  | {\tilde{A}} }}\ge t/2}} \\
			&\leq  2 \exp \Biggl( -c \min \Biggl(\frac{4t^2}{81C^2 K^4  \|\bd\|_{\infty}^4  m^2 ( 2\sqrt{n} + \sqrt{m})^4 },
			\frac{2t}{9C K^2 \| \bd\|_{\infty}^{2} ( 2\sqrt{n} + \sqrt{m})^2} \Biggr)\Biggr).
		\end{split}
	\end{align}
	Now note that 
	\begin{align}\label{eq:off:diag:on:diag}
		\begin{split}
			& \PP \pth{C\abs{\E\qth{\vv^\top  F \vv  | {\tilde{A}} } - m(m-1)\sum_{i=1}^n d_i^2}\ge t/2} \\
			&= \PP \pth{C\abs{\qth{  \sum_{i=1}^m \sum_{j \neq i, j=1}^m \tilde{\ba}_{j}^\top D^2 \tilde{\ba}_{j}} - m(m-1)\sum_{i=1}^n d_i^2}\ge t/2}  \\
			&\overset{(a)}{\leq} m \PP \pth{\abs{ \sum_{j\ne i_0, j=1}^m   \tilde{\ba}_{j}^\top D^2 \tilde{\ba}_{j} - (m-1)\sum_{i=1}^n d_i^2}\ge t/2m}  \\
			&\overset{(b)}{\leq} 2 m \cdot  e^{-c\min\left({t^2\over 4K^4m^2\nmhs{\mathbf{D}_{m-1}^2}^2 },{t  \over 2K^2m\nm{\mathbf{D}_{m-1}^2}_2 }\right)}  \\
			&{\leq} 2 m \cdot  e^{-c\min\left({t^2\over 4K^4 m^3n \|\bd\|_\infty^4 },{t  \over 2K^2m \|\bd\|_\infty^2 }\right)},
		\end{split}
	\end{align}
	where (a) follows using a union bound (the distribution for $\sum_{j\ne i_0, j=1}^m   \tilde{\ba}_{j}^\top D^2 \tilde{\ba}_{j}$ is the same for all $1\le i_0 \le m$), and (b) followed using \prettyref{lem:HW-ineq} for the $(m-1)n \times (m-1)n$ matrix $\mathbf D_{m-1}$ given by
	\begin{align*}
		\mathbf D_{m-1}^2 := \begin{bmatrix}
			D^2 & 0 & \cdots & 0 \\
			0 & D^2 & \cdots & 0 \\
			0 & 0 & \cdots & 0 \\
			0 & 0 & \cdots & D^2 \\
		\end{bmatrix}
	\end{align*}

	Finally, we get,
	\begin{align*}
		&\PP \Bigg(\bigg[  \sum_{i \ne j} |\av_{i}^\top  D \av_{j}|^2 -m(m-1) \sum_{i=1}^n d_i^2< -t \bigg] \cap  \calE_\xi \Bigg)\\
		&\le \PP \Bigg(\bigg[\bigg|  \sum_{i \ne j} |\av_{i}^\top  D \av_{j}|^2 -m(m-1) \sum_{i=1}^n d_i^2\bigg|> t \bigg] \cap  \calE_\xi \Bigg)\\
		&\le 2C  \exp \Biggl( -c \min \Biggl(\frac{4t^2}{81C^2K^4 \|\bd\|_{\infty}^4  m^2 (2\sqrt{n} + \sqrt{m})^4 },
		\frac{2t}{9CK^2 \| \bd\|_{\infty}^{2} ( 2\sqrt{n} + \sqrt{m})^2} \Biggr)\Biggr)  \nonumber \\
		&+ 2 m  \exp\left(-c\min\left({t^2\over 4K^4m^3n \|\bd\|_\infty^4 },{t  \over 2K^2m^2 \|\bd\|_\infty^2 }\right)\right)
		\nonumber\\
		&
		\leq 2m\tilde C \exp \left( -\tilde c \min \Big( \frac{t^2}{\tilde C^2   \| {\bd}\|_{\infty}^{{4}}  q_{m,n} }, \frac{t}{C \| {\bd}\|_{\infty}^{2} \tilde{q}_{m,n}} \Big)\right),
	\end{align*}
	for some constants $\tilde c$ and $\tilde C$ where $q_{m,n},\tilde q_{m,n}$ are as defined in the result statement. In view of \eqref{eq:m19} and the probability bound on $\calE_\xi$ via \prettyref{lem:singvalues}, this completes the proof.

}

\section{Likelihood function and its gradient} \label{app:MLE}
\subsection{Calculation of the likelihood function}
The aim of this section is to derive the log-likelihood for our model, 
\[
\yv_{\ell} = A X\wv_{\ell} + \zv_{\ell},  \ \ \ \ \ \ \ \  \ \ \ {\rm for}   \  \ \  \ell=1, \ldots, L, 
\]
where $\wv_1, \wv_2, \ldots, \wv_{L}$, and  $\zv_1, \zv_2, \ldots, \zv_L$ are independent and identically distributed $\mathcal{CN} (0, \sigma_w^2 I_n)$ and $\mathcal{CN} (0, \sigma_z^2 I_p)$ respectively. Since the noises are independent across the looks, we can write the log-likelihood for one of the looks, and then add the log-likelihoods to obtain the likelihood for all the looks. For notational simplicity, we write the measurements of one of the looks as: 
\[
\yv=AX\wv+\zv.
\]
Note that $\yv$ is a linear combination of two Gaussian random vectors and is hence Gaussian. Hence, by writing the real and imaginary parts of $\yv$ separately we will have
\[
\Re(\yv) + \Im(\yv) = (\Re(AX) + i \Im(AX))(\wv^{(1)} + i\wv^{(2)}) + (\zv^{(1)} + i\zv^{(2)}),
\]
and defining
\begin{align*}
\tilde{\yv} \triangleq \begin{bmatrix} \Re (\yv) \\ \Im (\yv) \end{bmatrix} &\sim \mathcal{N} \left( \begin{bmatrix} 0 \\ 0 \end{bmatrix},  B  \right),
\end{align*}
where
\begin{align*}
    B = \begin{bmatrix} \sigma_z^2 I_n + {\sigma_w^2}\Re (A X^2 \bar{A}^{\top}) & -\sigma_w^2 \Im (A X^2 \bar{A}^{\top}) \\ \sigma_w^2 \Im (A X^2 \bar{A}^{\top}) & \sigma_z^2 I_n + \sigma_w^2 \Re (A X^2 \bar{A}^{\top}) \end{bmatrix}.
\end{align*}

Hence, the log-likelihood of our data $\yv$ as a function of $\xv$ is
\begin{align}\label{eq:complex-likelihood-numerical}
    \ell(\xv) = &-\frac{1}{2} \log \det \left( B \right) \nonumber \\&- \frac{1}{2} 
    \begin{bmatrix}
        \Re (\yv^{\top}) & \Im (\yv^{\top})
    \end{bmatrix} \left(  B \right)^{-1} \begin{bmatrix} \Re (\yv) \\ \Im (\yv) \end{bmatrix} + C. 
\end{align}

Note that equation~\eqref{eq:complex-likelihood-numerical} is for a single look. Hence the log-likelihood of $\yv_1, \yv_2, \ldots, \yv_L$ as a function of $\xv$ is:
\begin{align*}
    \ell(\xv) = -\frac{L}{2}  \log \det(B) -\frac{1}{2} \sum_{\ell=1}^L
    \tilde{\yv}_\ell^{\top} B^{-1} \tilde{\yv}_\ell + C, 
\end{align*}
Since we would like to maximize $\ell(\xv)$ as a function of $\xv$, for notational simplicity we define the cost function $f_L(\xv): \mathbb R^n \to \mathbb R$:
\begin{align*}
    f_L(\xv) = \log \det (B) + \frac{1}{L } \sum_{\ell=1}^L \tilde{\yv}_\ell^{\top} B^{-1} \tilde{\yv}_\ell,
\end{align*}
that we will minimize to obtain the maximum likelihood estimate. 

\subsection{Calculation of the gradient of the likelihood function}

As discussed in the main text, to execute the projected gradient descent, it is necessary to compute the gradient of the negative log-likelihood function $\partial f_L$.
The derivatives of $f_L$ with respect to each element $\xv_j$ of $\xv$ is given by:
\begin{align}
    &\frac{\partial f_L}{\partial \xv_j} 
    = {2 \xv_j \sigma_w^2} \left( \begin{bmatrix} \Re (\av_{\cdot, j}^{\top}) 
    & \Im (\av_{\cdot, j}^{\top}) \end{bmatrix}
    B^{-1} 
    \begin{bmatrix} \Re (\av_{\cdot, j}) \\ \Im (\av_{\cdot, j}) \end{bmatrix} \right. 
    + \left.
    \begin{bmatrix} -\Im (\av_{\cdot, j}^{\top}) & \Re (\av_{\cdot, j}^{\top}) \end{bmatrix}
    B^{-1} 
    \begin{bmatrix} -\Im (\av_{\cdot, j}) \\ \Re (\av_{\cdot, j}) \end{bmatrix} \right) \nonumber \\
    &- \frac{2 \xv_j \sigma_w^2}{L } \sum_{\ell=1}^L \left[ \left(
    \begin{bmatrix} \Re (\av_{\cdot,j}^{\top}) & \Im (\av_{\cdot,j}^{\top}) \end{bmatrix} B^{-1}
    \begin{bmatrix} \Re (\yv_\ell) \\ \Im (\yv_\ell) \end{bmatrix} \right)^2 \right. 
    + \left.
    \left(
    \begin{bmatrix} -\Im (\av_{\cdot,j}^{\top}) & \Re (\av_{\cdot, j}^{\top}) \end{bmatrix}
    B^{-1} 
    \begin{bmatrix} \Re (\yv_\ell) \\ \Im (\yv_\ell) \end{bmatrix}
    \right)^2 \right] \nonumber \\
    &= 2 \xv_j \sigma_w^2 \left( \tilde{\av}_{\cdot, j}^{+T} B^{-1} \tilde{\av}_{\cdot, j}^{+} + \tilde{\av}_{\cdot, j}^{-T} B^{-1} \tilde{\av}_{\cdot, j}^{-} \right) 
    \quad - \frac{2 \xv_j \sigma_w^2}{L } \sum_{\ell=1}^L \left[ \left( \tilde{\av}_{\cdot, j}^{+T} B^{-1} \tilde{\yv}_\ell \right)^2 + \left( \tilde{\av}_{\cdot, j}^{-T} B^{-1} \tilde{\yv}_\ell \right)^2 \right], \label{eq:derivative}
\end{align}
where $\av_{\cdot,j}$ denotes the $j$-th column of matrix $A$, $\tilde{\av}_{\cdot, j}^+ = \begin{bmatrix}
    \Re (\av_{\cdot,j}) \\ \Im (\av_{\cdot, j})
\end{bmatrix}$ and $\tilde{\av}_{\cdot, j}^- = \begin{bmatrix}
    -\Im (\av_{\cdot, j}) \\ \Re (\av_{\cdot, j})
\end{bmatrix}$.

 \begin{algorithm}[t]
 	\caption{Bagged-DIP-based PGD algorithm}
 	\label{alg:PGD}
 	\begin{algorithmic}
 		\State {\bfseries Input:} $\{\yv_\ell\}^L_{\ell=1}, A, {\xv}_0 = \frac{1}{L} \sum^L_{\ell=1} |\bar{A}^{\top} \yv_\ell|, g_{\thetav}(\cdot), \delta_{\xv}$.
 		\State {\bfseries Output:} Reconstructed $\hat{{\bx}}$.
 		\For{$t=1,~\ldots, ~T$}
 		\State \textbf{[Gradient Descent Step]}
 		\If {$t=1$ \textbf{or} $\|\xv_{t}-\xv_{t-1}\|_{\infty} > \delta_{\xv}$}
 		\State Calculate exact $B^{-1}_{t} = (AX^2_{t}\bar{A}^{\top})^{-1}$.
 		\Else
 		\State \textbf{[Newton-Schulz matrix inverse approximation]}
 		\State $M^0 = B^{-1}_{t-1}$,
 		\For{$\tilde k=1,~\ldots, ~K$}
 		\State $M^{\tilde k} = M^{\tilde k-1} + M^{\tilde k-1} (I_m - B_t M^{\tilde k-1})$,
 		\EndFor
 		\State $\Tilde{B}^{-1}_{t}=M^K$.
 		\EndIf
 		\State Gradient calculation at coordinate $j$ as $\nabla f_L(\xv_{t-1,j})$ using $B^{-1}_{t}$ or $\Tilde{B}^{-1}_{t}$, 
        \State Update $\xv^G_{t,j}$: $\xv^{G}_{t,j} \leftarrow \xv_{t-1,j} - \mu_t \nabla f_L(\xv_{t-1,j})$. 
 		\State Save matrix inverse $B^{-1}_t$ or $\Tilde{B}^{-1}_{t}$.
 		\State Truncate $\xv^G_{t}$ into range $(0,1)$, $\xv^G_{t} = \text{clip}(\xv^G_{t}, 0, 1)$.
 		\State \textbf{[Bagged-DIP Projection Step]}
 		\State Generate random image given randomly generated noise $\uv \sim \mathcal{N}(\mathbf{0},I)$ as $g_{\thetav}(\uv)$.
 		\State Update $\thetav_t$ by optimizing over $\| g_{\thetav}(\uv) - \xv^G_t \|^2_2$ till converges: $ \hat{\thetav}_t \leftarrow \operatorname*{argmin}_{\thetav} \| g_{\thetav}(\uv) - \bx^G_t \|^2_2$.
 		\State Generate ${\bx}^P_t$ using trained $g_{\hat{\thetav}_{t}}(\cdot)$ as $\xv^P_t \leftarrow g_{\hat{\thetav}_{t}}(\uv)$.
 		\State Obtain $\xv_t = \xv^{P}_t $.
 		\EndFor
 		\State Reconstruct image as $\hat{{\bx}} = {\bx}_T$.
 	\end{algorithmic}
 \end{algorithm}

\subsection{Simplified form of the gradient} \label{ssec:matrixinversioncalc}
The special form of the matrix $B$ enables us to do the calculations more efficiently. To see this point, define:
\begin{align*}
    U + iV \triangleq \left( \sigma_z^2 I_n + {\sigma_w^2} A X^2 \bar{A}^{\top} \right)^{-1},
\end{align*}
where $U, V \in R^{m \times m}$. These two matrices should satisfy:
\begin{align*}
    \left( \sigma_z^2 I_n + {\sigma_w^2} \Re (A X^2 \bar{A}^{\top}) \right) U -{\sigma_w^2} \Im (A X^2 \bar{A}^{\top}) V = I_n \\
    {\sigma_w^2} \Im (A X^2 \bar{A}^{\top}) U + \left( \sigma_z^2 I_n + {\sigma_w^2}\Re (A X^2 \bar{A}^{\top}) \right) V = 0.
\end{align*}
These two equations imply that:
\begin{align}\label{eq:Bsymmetric}
    B^{-1} = \begin{bmatrix} U & -V \\ V & U \end{bmatrix}.
\end{align}

This simple observation, enables us to reduce the number of multiplications required for the Newton-Schulz algorithm. More specifically, instead of requiring to multiply two $2m \times 2m$ matrices, we can do $4$ multiplications of $m \times m$ matrices. This helps us have a factor of $2$ reduction in the cost of matrix-matrix multiplication in our Newton-Schulz algorithm. 

In cases the exact inverse calculation is required, again this property enables us to reduce the inversion of matrix $B \in \mathbb{R}^{2m \times 2m}$ to the inversion of two $m \times m$ matrices (albeit a few $m\times m$ matrix multiplications are required as well). 

Plugging \eqref{eq:Bsymmetric} into~\eqref{eq:derivative}, we obtain a simplified form for the gradient of $f_L(\xv)$:
\begin{align*}
    \frac{\partial f_L}{\partial \xv_j} =& {4 \xv_j \sigma_w^2}  \Re\left(\bar{\av}_{\cdot, j}^{\top} (U + iV) \av_{\cdot, j}\right)
    - \frac{2 \xv_j \sigma_w^2}{L} \sum_{\ell=1}^L \big[ \Re^2 \left( \bar{\av}_{\cdot, j}^{\top} (U + iV) \yv_\ell \right) 
    \notag\\
    & + \Im^2 \left( \bar{\av}_{\cdot, j}^{\top} (U + iV) \yv_\ell \right) \big] \nonumber \\
    =& {4 \xv_j \sigma_w^2} \Re\left(\bar{\av}_{\cdot, j}^{\top} (U + iV) \av_{\cdot, j} \right) 
    - \frac{2 \xv_j \sigma_w^2}{L } \sum_{\ell=1}^L \left| \bar{\av}_{\cdot, j}^{\top} (U + iV) \yv_\ell \right|^2.
\end{align*}

 \begin{figure*}[t]
 	\centering
 	\includegraphics[width=0.99\textwidth]{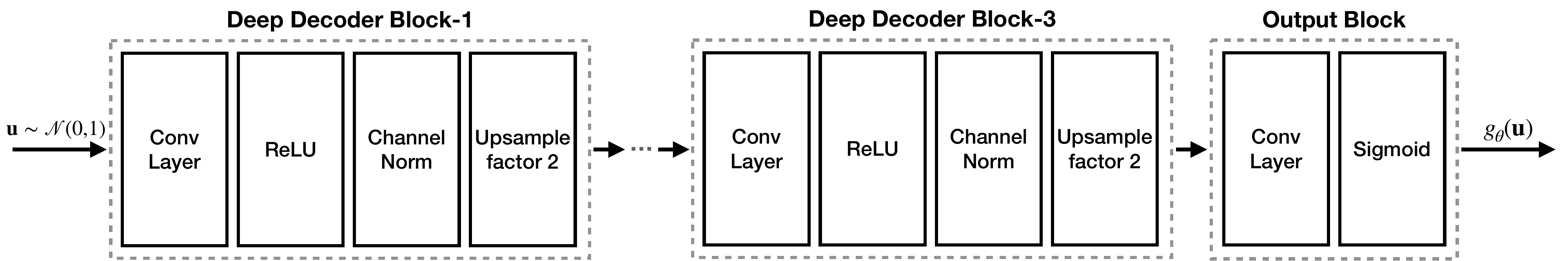}
 	\caption{We adopt the Deep Decoder architecture, a simplified variant of DIP, as the building block of our networks. For consistency and to avoid confusion, we refer to it as DIP throughout the paper.}
 	\label{fig:DIP_block_structure}
 \end{figure*}

\subsection{Details of the Bagged-DIP-based PGD algorithm}
\label{app:add_experiment}
 Algorithm~\ref{alg:PGD} shows a detailed version of the final algorithm we execute for recovering images from their multilook, speckle-corrupted, undersampled measurements. In one of the steps of the algorithm, we ensure that all the pixel values of our estimate are within the range $[0,1]$. This is because we have assumed that the image pixels take values within $[0,1]$.

The only remaining parts of the algorithm we need to clarify are (1) our hyperparameter choices and (2) the implementation details of the Bagged-DIP module. As described in the main text, in each (outer) iteration of PGD, we learn three DIPs and then take the average of their outputs. Let us now consider one of these DIPs that is applied to one of the $h_k \times w_k$ patches. 
 
 Inspired by the network structure of Deep Decoder \cite{heckel2018deep}, we construct our neural network using four blocks: we call the first three blocks Deep Decoder blocks and the last one output block. 
 The structures of the blocks are shown in Figure~\ref{fig:DIP_block_structure}. As is clear from the figure, each Deep Decoder block is composed of the following components:
 \begin{itemize}
 	\item Up-sample: This unit increases the height and width of the datacube that receives by a factor of 2. To interpolate the missing elements, it uses the simple bilinear interpolation. Hence, if the size of the image is $128 \times 128$, then the height and width of the input to Deep Decoder Block-3 will be $64 \times 64$, the input of Deep Decoder Block-2 will be $32 \times 32$, and so on.
 	\item ReLU: this module is quite standard and does not require further explanation. 
 	
 	\item Convolution: For all our simulations, we have either used $1\times 1$ or $3 \times 3$ convolutions. Additionally, we provide details on the number of channels for the data cubes entering each block in our simulations. The channel numbers are [128, 128, 128, 128] for the four blocks.
 	
 \end{itemize}
 
The output block is simpler than the other three blocks. It only has a 2D convolution that uses the same size as the convolutions of the other Deep Decoder blocks. The nonlinearity used here is a sigmoid function, as we assume that the pixel values are between $[0,1]$.

Finally, we should mention that each element of the input noise $\uv$ of DIP (as described before DIP function is $g_{\thetav} (\uv)$) is generated independently from Normal distribution $\mathcal{N}(0,1)$.

 The other hyperparameters that are used in the DIP-based PGD algorithm are set in the following way: The learning rate of the loglikelihood gradient descent step (in PGD) is set to $\mu=0.001$ for $L \leq 8$, and $0.01$ otherwise. For training the Bagged-DIP, we use Adam~\cite{kingma2014adam} with the learning rate set to $0.001$ and weight decay set to $0$. The number of iterations used for training Bagged-DIP for different estimates on images are provided in Table~\ref{tab:num_iteration_DIP}, the numbers are doubled when $m/n=1.0$. We run the outer loop (gradient descent of likelihood) for $100,200,300$ iterations when $m/n=0.5,0.25,0.125$ respectively. For ``Cameraman" only, when $m/n=0.125$, since the convergence rate is slow, we run $800$ outer iterations.
 
 The Newton-Schulz algorithm, utilized for approximating the inverse of the matrix \( B_t \), has a quadratic convergence when the maximum singular value \( \sigma_{\max}(I - M^0 B_{t}) < 1 \). Hence, ideally, if this condition does not hold, we do not want to use the Newton-Schulz algorithm and may prefer the exact inversion. Unfortunately, checking the condition \( \sigma_{\max}(I - M^0 B_{t}) < 1 \) is also computationally demanding. However, the special form of $B_t$ enables us to have an easier heuristic evaluation of this condition.

\begin{figure}[ht]
	\centering
	\includegraphics[width=0.47\linewidth]{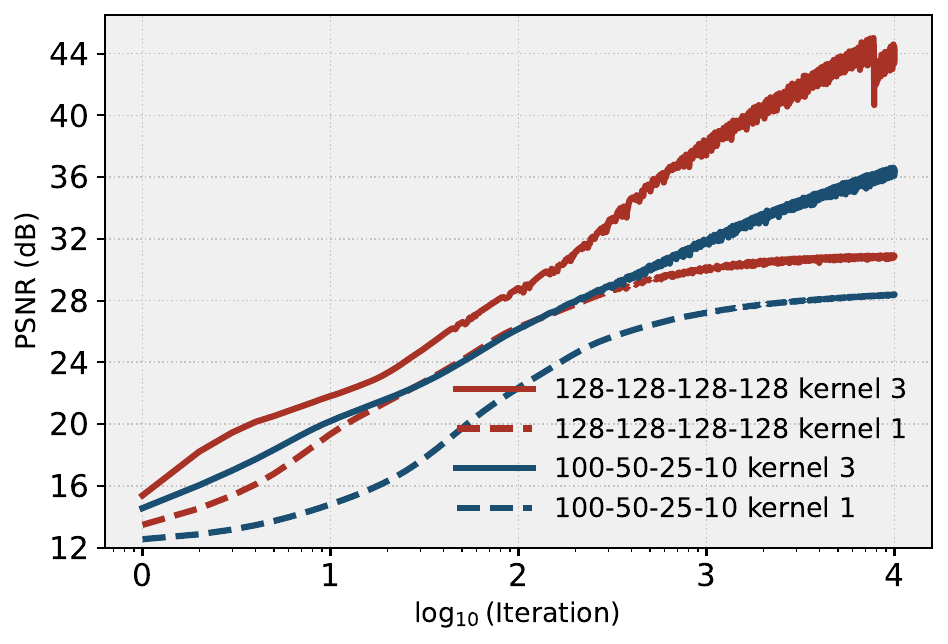}
	\includegraphics[width=0.47\linewidth]{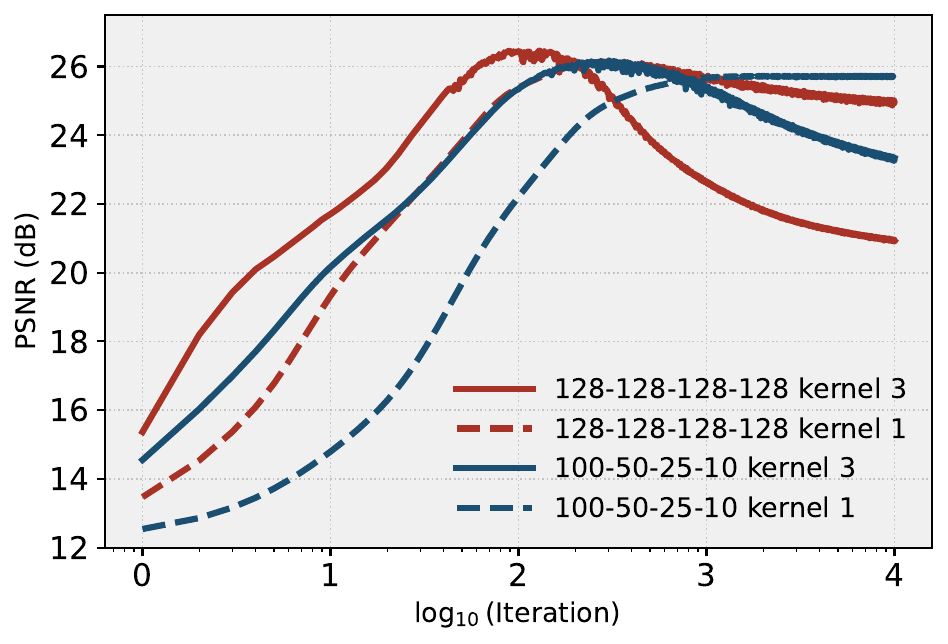}
	
	\caption{PSNR (averaged over 8 images) versus iteration count is depicted for four DIP models fitted to both clean (Left) and noisy images with only additive noise, noise level $\sigma=25$ (Right). The 4-layer networks used in DIP are specified in the legend.}
	\label{fig:decoder_compare_clean}
\end{figure}

\begin{table*}[t]
    \centering
    \caption{Number of iterations used in training Bagged-DIP for different estimates.}
    \label{tab:num_iteration_DIP}
    \vspace{0.5em}
    \begin{tabular}{@{}llllllllll@{}}
        \toprule
        & \textbf{Patch size} & \textbf{Barbara} & \textbf{Peppers} & \textbf{House} & \textbf{Foreman} & \textbf{Boats} & \textbf{Parrots} & \textbf{Cameraman} & \textbf{Monarch} \\
        \midrule
        & 128 & 400 & 400 & 400 & 400 & 400 & 800 & 4000 & 800 \\
        & 64  & 300 & 300 & 300 & 300 & 300 & 600 & 2000 & 600 \\
        & 32  & 200 & 200 & 200 & 200 & 200 & 400 & 1000 & 400 \\
        \bottomrule
    \end{tabular}
    \vspace{-0.1in}
\end{table*}
 
 For our problems, we establish an empirical sufficient condition for convergence: \( \|\xv_t - \xv_{t-1}\|_{\infty} < \delta_{\xv} \), where \( \delta_{\xv} \) is a predetermined constant. To determine the most robust value for \( \delta_{\xv} \), we conducted simple experiments. We set \( n = 128 \times 128 \) and \( m / n = 0.5 \). The sensing matrix \( A \) is generated as described in the main part of the paper (see Section \ref{sec:simulation_baggedDIP}). Each element of \( {\xv}_o \) is independently drawn from a uniform distribution \( U[0.001, 1] \). Furthermore, each element of \( \Delta {\xv}_o \) is independently sampled from a two-point distribution. In this distribution, the probability of the variable \( X \) being \( \delta_{\xv} \) is equal to the probability of \( X \) being \( -\delta_{\xv} \), both with a probability of 0.5, ensuring \( \|\Delta \xv_o\|_{\infty} = \delta_{\xv} \). We define \( B \) as \( A (X + \Delta X_o)^2 \bar{A}^{\top} \), and \( M^0 \) as \( (A X^2 \bar{A}^{\top})^{-1} \). We then assess the convergence of the Newton-Schulz algorithm for calculating \( B^{-1} \). For various values of \( \delta_{\xv} \), we ran the simulation 100 times each, recording the convergence success rate. As indicated in Table \ref{tab:threshold}, the algorithm demonstrates instability when \( \delta_{\xv} \geq 0.13 \). Consequently, we set \( \delta_{\xv} \) to 0.12 in all our simulations to ensure the reliable convergence of the Newton-Schulz algorithm.

\begin{table}[ht]
    \centering
    \caption{Convergence success rate under varying threshold $\delta_{\xv}$ in Newton-Schulz algorithm.}
    \label{tab:threshold}
    \begin{tabular}{c c}
        \toprule
        \textbf{Threshold} $\delta_{\xv}$ & \textbf{Convergence Success Rate} \\
        \midrule
        0.10 & 100\% \\
        0.11 & 100\% \\
        0.12 & 100\% \\
        0.13 & 38\% \\
        0.14 & 0\% \\
        0.15 & 0\% \\
        \bottomrule
    \end{tabular}
\end{table}

\subsection{Study on general properties of DIP}
\label{app:DIP}

Fig.~\ref{fig:decoder_compare_clean} shows that higher-capacity DIP networks accurately fit clean images but tend to overfit in the presence of noise, whereas simpler architectures, while less expressive, exhibit greater robustness to noise. This trade-off between expressivity and robustness is consistent with prior observations~\cite{heckel2018deep, heckel2019denoising}.

{
\subsection{Empirical upper bound experimental setting}
\label{app:empirical_UB}
We empirically study the empirical upper bound of the problem, i.e., recovering images from multiple looks of undersampled measurements. We design the experiment as follows: The DnCNN is trained to estimate $\xv$ from the networks' input $\sqrt{\frac{1}{L} \sum_{\ell=1}^{L} |\bar{A}^{\top} \yv_\ell|^2}$ using supervised learning. All images are loaded in grayscale and normalized to the range $[0,1]$ The training data is augmented from the BSD400 image dataset. 16 smaller images with the size of $128 \times 128$ are randomly cropped from the raw image with larger size ($481 \times 321$ or $321 \times 481$) in the dataset. The measurement is generated as $\yv_\ell = A X_o \wv_\ell + \zv_\ell$ from the clean image $\xv_o$, where $m/n=1.0$ in $A$, and $\sigma_z=25$ in $z$.
The model is trained for 100 epochs using the Adam optimizer with learning rate $10^{-3}$, batch size 16, and no weight decay. A StepLR scheduler is employed to decay the learning rate by a factor of 0.5 every 30 epochs. The training objective is the mean squared error (MSE). We report performance using PSNR and SSIM. The best model is selected based on the highest achievable testing PSNR.
The DnCNN network consists of 17 convolutional layers with kernel size $3 \times 3$ and 64 feature channels in the hidden layers. 
All convolutions use padding of 1 to preserve spatial resolution. The final output is passed through a sigmoid activation, which constrains the reconstructed image to the range $[0,1]$.
Unlike the original DnCNN formulation that predicts residual noise, our model directly regresses the clean image from the corrupted input.
}

{
\subsection{Visualization of the reconstructed images}
\label{app:visualization}

We present the visualization to better show the reconstructed images under different system's parameters, including sampling rates $m/n=0.125, 0.25, 0.5$ and numbers of looks $L=1,2,4,8,16,32,64,128$.
}

\begin{figure}[H]
    \centering
\includegraphics[width=\linewidth]{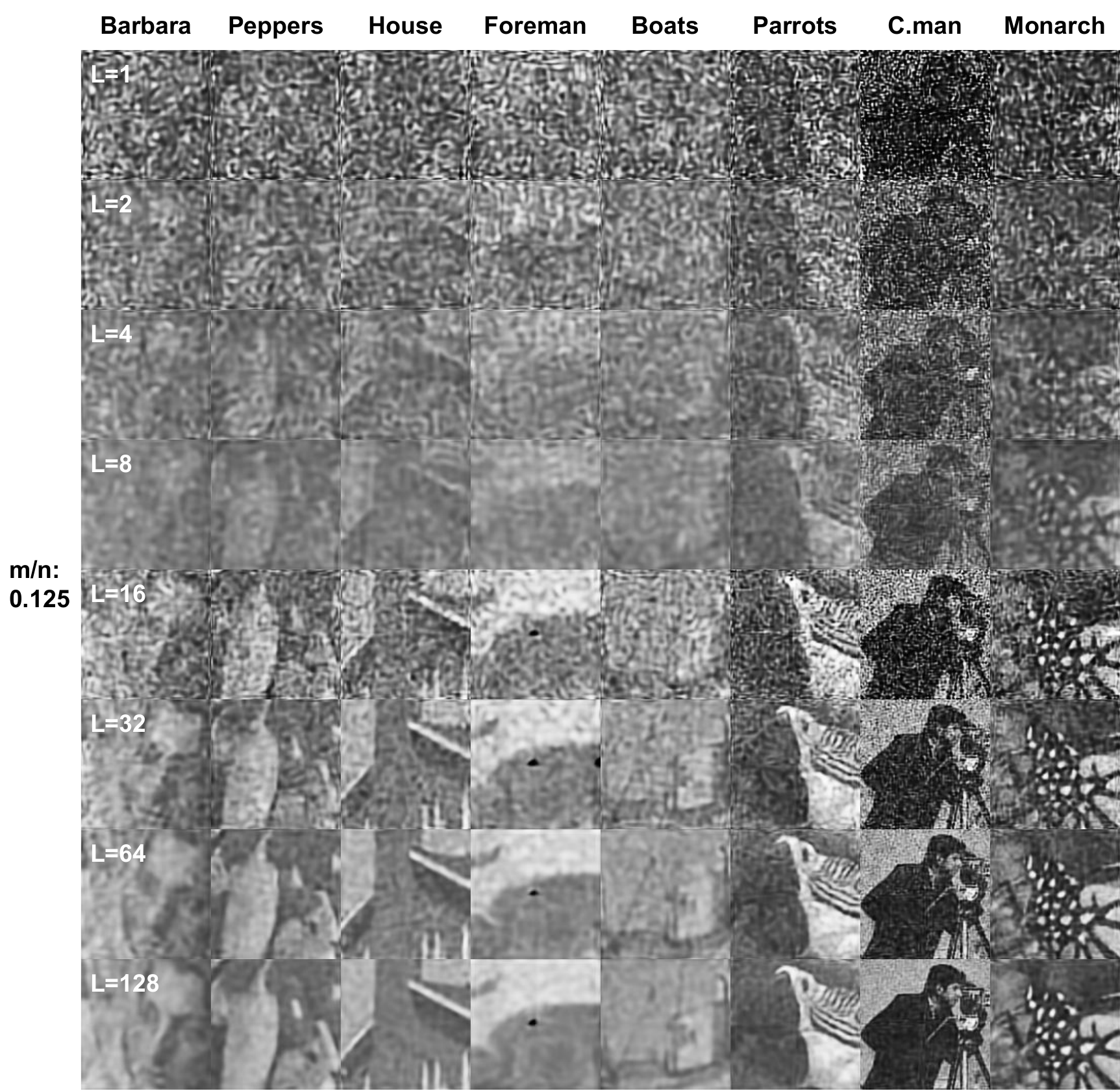}
    \caption{8 reconstructed images from $L=1,2,4,8,16,32,64,128$ looks of $12.5\%$ downsampled complex-valued measurements.}
    \label{fig:vis_S125}
\end{figure}

\begin{figure}[H]
    \centering
\includegraphics[width=\linewidth]{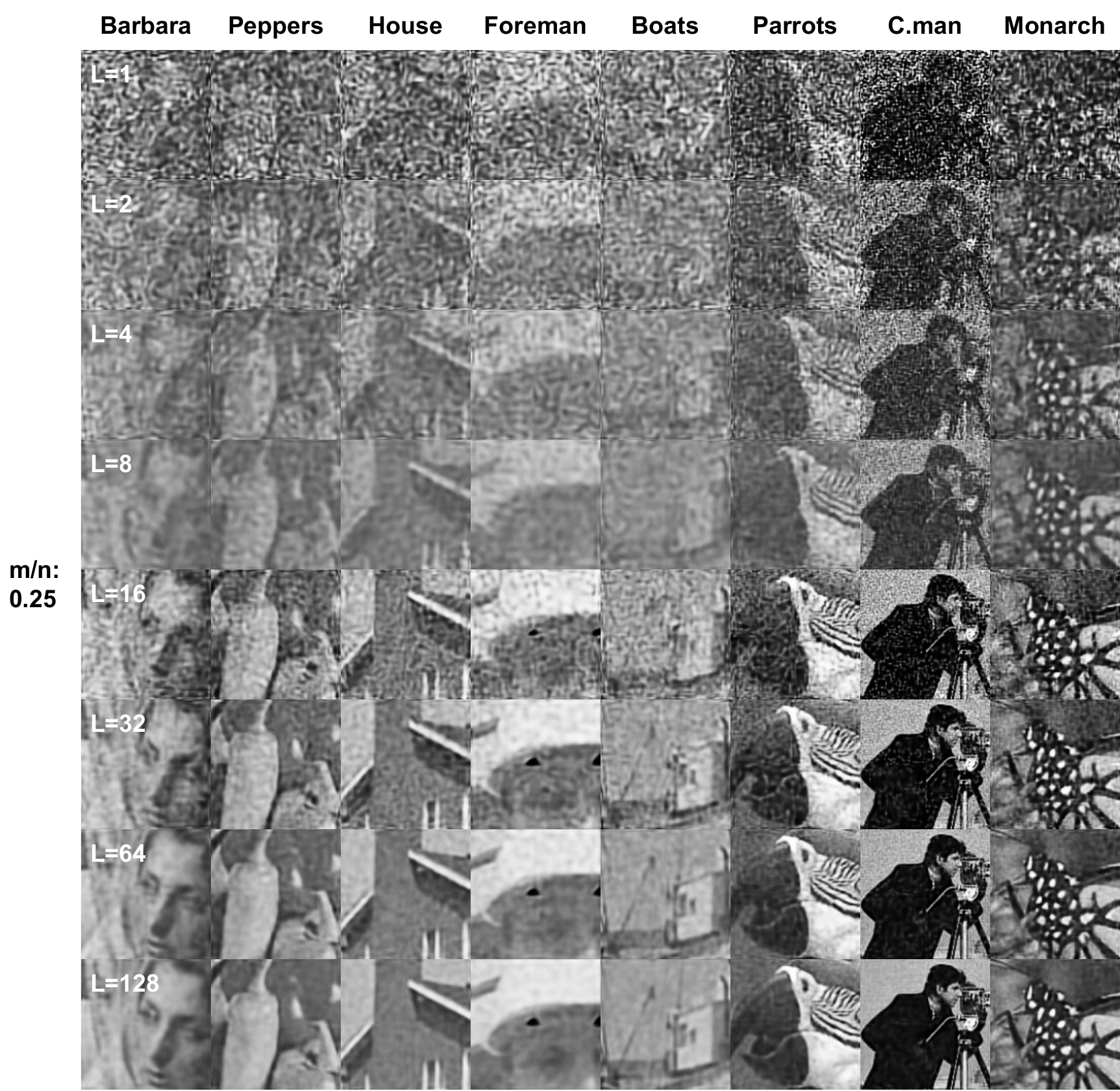}
    \caption{8 reconstructed images from $L=1,2,4,8,16,32,64,128$ looks of $25\%$ downsampled complex-valued measurements.}
    \label{fig:vis_S25}
\end{figure}

\begin{figure}[H]
    \centering
\includegraphics[width=\linewidth]{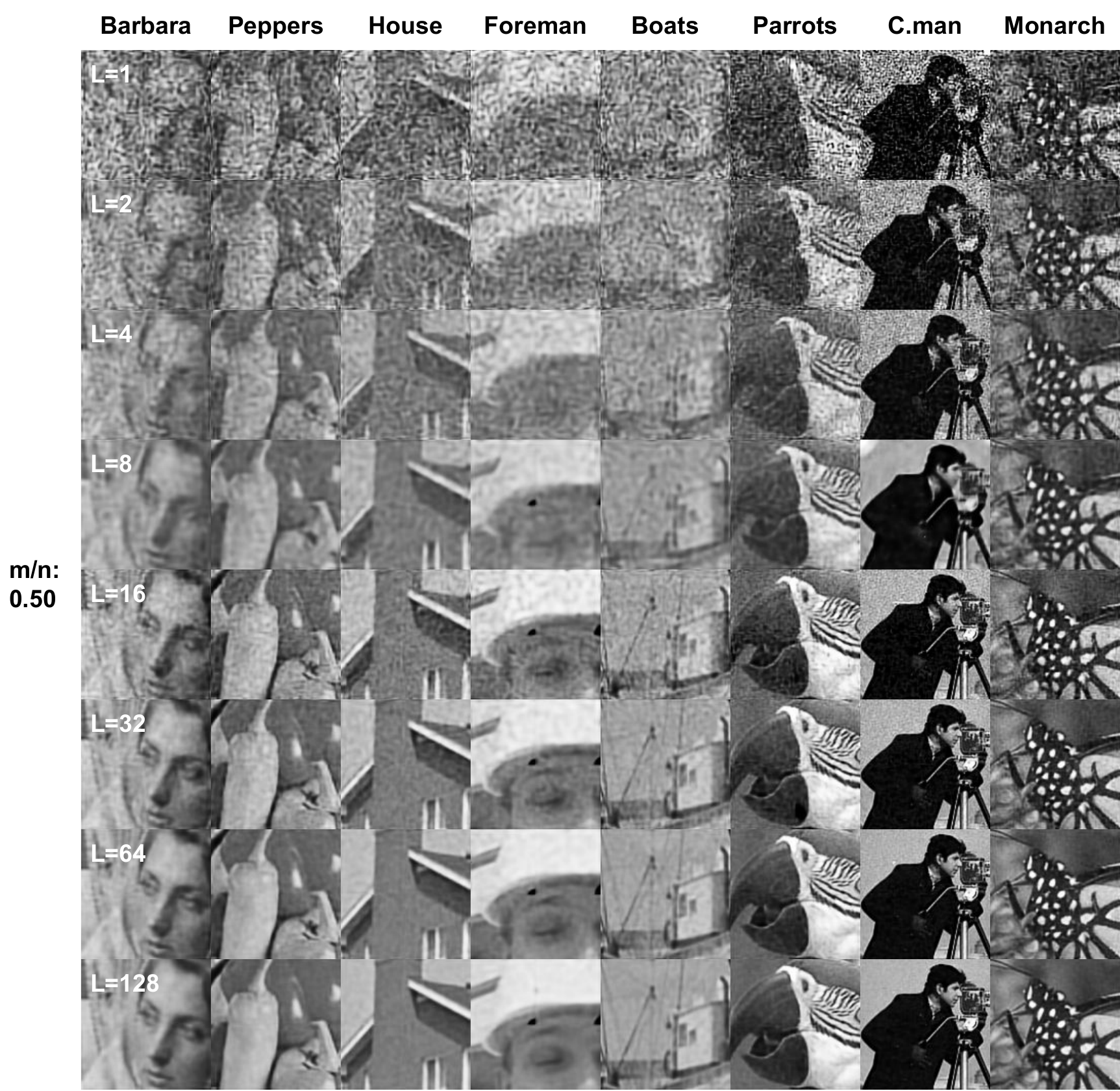}
    \caption{8 reconstructed images from $L=1,2,4,8,16,32,64,128$ looks of $50\%$ downsampled complex-valued measurements.}
    \label{fig:vis_S50}
\end{figure}

\end{document}